\documentclass[11pt,a4paper]{article}

\usepackage[utf8]{inputenc}
\usepackage[a4paper, margin=2cm]{geometry}
\usepackage{amsmath}
\usepackage{amsmath, amssymb}
\usepackage{graphicx}
\usepackage{wrapfig} 
\usepackage{subcaption}
\usepackage{booktabs}
\usepackage{float}
\usepackage{algorithm}
\usepackage{algorithmic}
\usepackage[numbers]{natbib}
\usepackage{parskip}
\usepackage{makecell}
\usepackage{multirow}
\usepackage{url}
\usepackage{rotating}
\usepackage{caption}
\usepackage{IEEEtrantools}
\usepackage[aboveskip=1pt,labelfont=bf,labelsep=period,justification=raggedright,singlelinecheck=off]{caption}

\title{\textbf{A Novel Deep Hybrid Framework with Dynamic Multi-Objective Fitness Optimization for Robust Real-Time Human Activity Recognition}}

\author{
    Wasi Ullah$^1$, 
    Yasir Noman Khalid$^1$,\\
    Alanoud S.Al Mazroa$^2$,
    Obada Al-Khatib$^3$,
    Julian Hoxha$^4$,
    Saddam Hussain Khan$^5$
    \\
    \\
    \small $^1$Department of Computer Science, HITEC University, Museum Road, Taxila, Pakistan \\
    \small $^2$Department of Information Systems,\\ College of Computer and Information Sciences, \\Princess Nourah bint Abdulrahman University (PNU), \\P.O. Box 84428, Riyadh 11671, Saudi Arabia. \\
    \small $^3$University of Wollongong, Dubai, UAE. \\
    \small $^4$College of Engineering and Technology, American University of Middle East, Egaila, Kuwait. \\
    \small $^5$Department of Computer Systems Engineering, UEAS, Swat, Pakistan \\
    \small Email: wasi.ullah@student.hitecuni.edu.pk
}

\date{}

\begin{document}

\maketitle

% Optional DOI or reference
% \noindent\textbf{DOI:} 10.1109/ACCESS.2017.DOI
\section*{Abstract}
Real-time Human Activity Recognition (HAR) has wide-ranging applications in areas such as context-aware environments, public safety, assistive technologies, and autonomous monitoring and surveillance systems. However, existing real-time HAR systems face significant challenges, including limited scalability and high computational costs arising from redundant features. To address these issues, the Inception-V3 model was customized with region-based and boundary-aware operations, using average pooling and max pooling, respectively, to enhance region homogeneity, suppress noise, and capture discriminative local features, while improving robustness through down-sampling. Furthermore, to effectively encode motion dynamics, an Attention-Augmented Long Short-Term Memory (AA-LSTM) network was employed to learn temporal dependencies across video frames. Features are extracted from video dataset and are then optimized through a novel proposed dynamic composite feature selection method called Adaptive Dynamic Fitness Sharing and Attention (ADFSA). This ADFSA mechanism is embedded within a genetic algorithm to select a compact, optimized subset of features by dynamically balancing multiple objectives, accuracy, redundancy reduction, feature uniqueness, and complexity minimization. As a result, the selected subset of diverse and discriminative features enables lightweight machine learning classifiers to achieve accurate and robust HAR in heterogeneous environments. Experimental results demonstrate up to 99.65\% accuracy using as few as seven selected features, with improved inference time on the challenging UCF-YouTube dataset, which includes factors such as occlusion, cluttered backgrounds, complex motion dynamics, and poor illumination conditions.

\textbf{Keywords:} Human Activity Recognition, Hybrid Deep Learning, InceptionV3, LSTM Networks, Dynamic Feature Selection, Adaptive Dynamic Fitness Sharing, Real-Time Processing, Edge Computing

\section{Introduction}

Human activity recognition (HAR) plays a pivotal role in the development of smart cities, particularly in the context of video-based surveillance. Accurate identification of human behavior can help law enforcement and public safety agencies detect irregular activities ranging from accidents and property damage to illicit behaviors such as fighting and theft. Early research in HAR focused primarily on controlled environments with a single actor, but recent studies have shifted toward more realistic and challenging scenarios. These include uncontrolled video datasets with complex backgrounds, varying illumination, camera movement, and occlusion, making accurate recognition considerably more difficult \cite{b1}. Improving pedestrian detection and tracking in such dynamic surveillance situations remains a significant challenge.

Although Convolutional Neural Networks (CNNs) have demonstrated remarkable performance in image classification and object detection, they are inherently limited to processing individual frames. They cannot directly capture temporal dynamics in video sequences \cite{b2}. To address this, two-stream CNNs were presented, in which appearance and motion information (e.g., optical flow) are processed separately. Although they achieve good recognition performance, they incur high computational costs, limiting their suitability for large-scale or real-time applications \cite{b3,b4}. More recently, 3D CNNs have been introduced to directly extract spatiotemporal information from raw video streams \cite{b5, b6}. To enhance temporal modeling, recurrent architectures such as RNNs and LSTMs have also been incorporated \cite{b7,b8}. These approaches suffer from high computational cost, especially when processing long video sequences, which limits their applicability for large-scale and real-time scenarios.

Recently, Transformer-based models have become popular in human activity recognition. For video data, researchers have combined 3D convolution with attention-based Transformers to capture spatial and temporal patterns, while efficiency-focused architectures have enabled lightweight, real-time processing in open-set HAR tasks \cite{b9,b10}. In the wearable sensor domain, methods such as MoPFormer, SETransformer, and P2LHAP have introduced novel innovations, including tokenization in motion primitives, channel-wise attention, and sequence forecasting, achieving improved accuracy, interpretability, and generalization across datasets \cite{b11,b12}. Lightweight, end-to-end Transformers for smartphone IMU data also match or exceed CNN–LSTM baselines with fewer parameters and strong performance \cite{b13}. Despite these gains, recent surveys note that Transformer models often entail heavy computation, dependency on large datasets, and limitations when deployed on resource-constrained devices, posing challenges to their broader adoption \cite{b14}.

To address these limitations, researchers have explored feature fusion and selection strategies for HAR. Promising performance achieved by incorporating deep spatial and motion representations using spatial–temporal attention mechanisms \cite{b15} and combining hand-crafted multiview features with feature selection (FS) based on entropy variability \cite{b16}. They often rely on hand-crafted components and produce high-dimensional feature representations, leading to increased computational complexity and limited scalability. Efficient selection of discriminative features is therefore essential to improve inference speed without compromising accuracy.

In this work, we propose a novel hybrid HAR framework that integrates a customized InceptionV3 CNN for spatial feature extraction with an Attention-Augmented Long-Short-term Memory Network (AA-LSTM) for temporal modeling.  CNN captures discriminative spatial patterns from individual frames, while AA-LSTM models temporal dependencies across sequences. Designed for real-time video processing, the framework handles challenges such as occlusion, cluttered backgrounds, motion blur, and multiview variations, as exemplified by the UCF YouTube action dataset \cite{b17}. Both models are end-to-end trainable and optimized via transfer learning and fine-tuning to ensure efficient inference. Furthermore, to reduce feature redundancy and computational overhead, a genetic algorithm (GA)–based FS strategy is applied, selecting only the most discriminative spatiotemporal features for final classification. By eliminating the need for costly preprocessing steps such as optical flow or skeleton extraction, the proposed framework achieves high accuracy while maintaining practical real-time applicability.

The following are the main contributions of this work:
\begin{itemize}
    \item This study proposes a hybrid DL framework that synergistically integrates spatial customized InceptionV3 and temporal AA-LSTM modeling for HAR from video data.  The customized InceptionV3 integrates region-based (average pooling) and boundary-aware (max pooling) operations after each convolutional block to enhance region homogeneity, suppress noise, and capture discriminative local features, while improving robustness through down-sampling. Moreover, the enhanced InceptionV3 extracts fine-grained localization, contextual cues, and human positioning within frames, which an AA-LSTM sequentially processes to model inter-frame temporal dependencies and motion dynamics with high fidelity.
    \item  A novel GA-driven FS strategy based on a dynamic composite fitness function, ADFSA, is employed to achieve compact and discriminative feature subsets. The ADFSA mechanism dynamically reweights multiple objectives: classification accuracy, redundancy minimization, uniqueness enhancement, and complexity reduction through attention-guided adaptation, thereby accelerating convergence and mitigating local stagnation during the evolutionary process.
    \item The optimally selected and diverse feature subsets include both spatial and temporal, are highly compact, diverse, and discriminative, enabling lightweight ML classifiers to achieve robust HAR performance across heterogeneous conditions with minimal resource overhead.
    \item The proposed framework outperforms state-of-the-art approaches on the UCF–YouTube dataset with 99.65\% recognition accuracy, while reducing the feature space from 128 to 7. This compact representation enables real-time inference on edge devices such as a Raspberry Pi.
.

\end{itemize}
 The rest of the paper is organized as follows. Section II provides related work. Section III presents the methodology and proposes the ADFSA scheme. Section IV is related to the experimental setup. Results and discussion are elaborated in section V. Section VI concludes the paper.

\section{Related Work}

Recent research has increasingly emphasized advanced DL architectures for HAR, spanning CNN-based frameworks, transfer learning paradigms, and, more recently, Transformer-driven sequence modeling. The BT-LSTM model~\cite{b18}, for instance, achieved 85.3\% accuracy on the UCF11 dataset, highlighting the importance of temporal modeling but exhibiting limited generalization in unconstrained scenarios. To jointly capture spatial and temporal dependencies, several end-to-end architectures have been explored. The CNN--BiLSTM framework~\cite{b19} attained 92.84\% accuracy, while the deep autoencoder--CNN approach~\cite{b20} reached 96.2\%. However, these models required large-scale training data and involved substantial computational costs, reducing their feasibility in real-time applications.  

Efforts to strengthen spatial representation also included handcrafted and hybrid descriptors. The key-frame driven KFDI method~\cite{b21} achieved only 79.4\%, largely due to weak spatial discrimination. More sophisticated designs, such as the dilated CNN--BiLSTM with residual blocks~\cite{b22} and the local--global feature fusion with QSVM~\cite{b23}, achieved 89.01\% and 82.6\%, respectively. However, these methods exhibited reduced robustness when applied to complex activities with occlusions, cluttered backgrounds, and motion variations.  

More recently, fully 3D CNN and Transformer-based encoders have been introduced to address long-term temporal modeling. The 3D-CNN~\cite{b24} reported 85.20\% accuracy, whereas the ConvNeXt--TCN framework~\cite{b25} achieved 97.73\%, underscoring the role of temporal aggregation. Hybrid architectures such as VGG--BiGRU~\cite{b26} and ViT-ReT~\cite{b27} attained 93.38\% and 92.4\% accuracy, respectively. However, despite these advancements, most of these models employed parameter-heavy architectures and retained high-dimensional feature representations, resulting in increased inference latency and limited applicability on resource-constrained platforms.  

Additionally, the bidirectional LSTM model~\cite{b28} reported a high recognition accuracy of 99.2\%, yet it relied on full feature vectors without any dimensionality reduction. This dependence on high-dimensional features limited its scalability and efficiency in real-time or edge-deployment environments. Overall, prior studies have demonstrated significant progress but continue to face challenges related to computational efficiency, redundancy in feature representation, and deployment feasibility. A concise summary of existing approaches is presented in Table~\ref{tab:har_comparison}.

\begin{sidewaystable}[!htp]
\centering
\caption{Summary of Previous Works on Human Activity Recognition}
\begin{tabular}{|c|c|c|p{5.8cm}|}
\hline
\textbf{Ref.} & \textbf{Dataset(s)} & \textbf{Results / Method} & \textbf{Key Shortcomings} \\ \hline
\cite{b18} & UCF11, KTH & Block-term tensor RNN; 94\% (UCF11) & Requires tensor decomposition + full RNN; high computation \\ \hline
\cite{b19} & UCF11, KTH & CNN features + BiLSTM; 94.9\% & No feature reduction → high-dimensional inputs → slower inference \\ \hline
\cite{b20} & UCF11, surveillance & Deep autoencoder + CNN; 93.6\% & Lacks explicit temporal modeling; not optimized for real-time \\ \hline
\cite{b21} & UCF11 & Dynamic Image + CNN; 95.7\% & Ignores temporal sequences → limited spatio-temporal modeling \\ \hline
\cite{b22} & UCF11, HMDB51, UCF50 & Dilated CNN + Attention-LSTM; 96.5\% (UCF11) & Heavy attention block; poor generalization (68.4\% HMDB51) \\ \hline
\cite{b23} & EUCF-11 / RUCF-11 & Spiking CNN for event data; 92.5\% / 95.7\% & Designed for neuromorphic data, not RGB video; no feature reduction \\ \hline
\cite{b24} & UCF101 & 3D-CNN; 94.1\% & High memory and inference time due to 3D convolutions \\ \hline
\cite{b25} & UCF11, UCF50, UCF101, JHMDB & ConvNeXt + TCN; 97.73\%, 98.81\%, 98.46\%, 83.38\% & Very accurate, but deep/heavy model; no feature compression \\ \hline
\cite{b26} & UCF Sports, JHMDB, UCF101 & CNN + BiGRU + FS; 95.2\%, 78.7\%, 93.4\% & Features still high-dimensional; no real-time deployment \\ \hline
\cite{b27} & UCF101, HMDB51 & ViT-ReT; 97.4\%, 82.6\% & Transformer complexity; high inference cost \\ \hline
\cite{b28} & UCF11, JHMDB & TL + BiLSTM; 96.7\%, 79.5\% & Uses full feature vectors; no dimensionality reduction \\ \hline
\end{tabular}
\label{tab:har_comparison}
\end{sidewaystable}

Despite notable progress, existing HAR models still face challenges related to computational efficiency, redundancy in feature representation, and deployment feasibility on real-time or edge platforms. This emphasizes the necessity for frameworks that are both lightweight and efficient while maintaining high recognition accuracy and scalability. To address these challenges, this research introduces a new deep hybrid framework that incorporates a dynamic composite ADFSA FS approach. This technique combines various optimization objectives such as accuracy, redundancy, uniqueness, and complexity, which adapt their importance levels during the evolutionary process. Moreover, by utilizing a dual-stage deep feature extraction process (custom InceptionV3 + AA-LSTM), the method effectively captures spatial and temporal features. The hybrid structure of this framework allows for improved recognition performance, quicker inference, and decreased model complexity, making it suitable for real-time Human Activity Recognition (HAR) applications.

\section{Methodology}

This study presents a novel hybrid deep feature extraction and selection framework for HAR, structured into four sequential phases:  

\begin{enumerate}
    \item \textbf{Data Preprocessing:} Video data is decomposed into frames, uniformly sampled, resized to \(224 \times 224\) pixels, normalized, and augmented through controlled transformations to enhance robustness and mitigate overfitting.  

    \item \textbf{Spatial Feature Extraction:} A customized InceptionV3 captures fine-grained spatial descriptors, including contextual cues, region homogeneity, and discriminative local patterns.  

    \item \textbf{Temporal Modeling:} An AA-LSTM encodes inter-frame dependencies and motion dynamics with high temporal fidelity.  

    \item \textbf{Feature Selection and Classification:} A dynamic composite fitness function-based genetic algorithm with ADFSA identifies compact, non-redundant, and highly discriminative feature subsets, which are subsequently classified using lightweight machine learning models.  
\end{enumerate}

\begin{figure}[!h]
    \centering
    \includegraphics[width = 1.0 \textwidth]{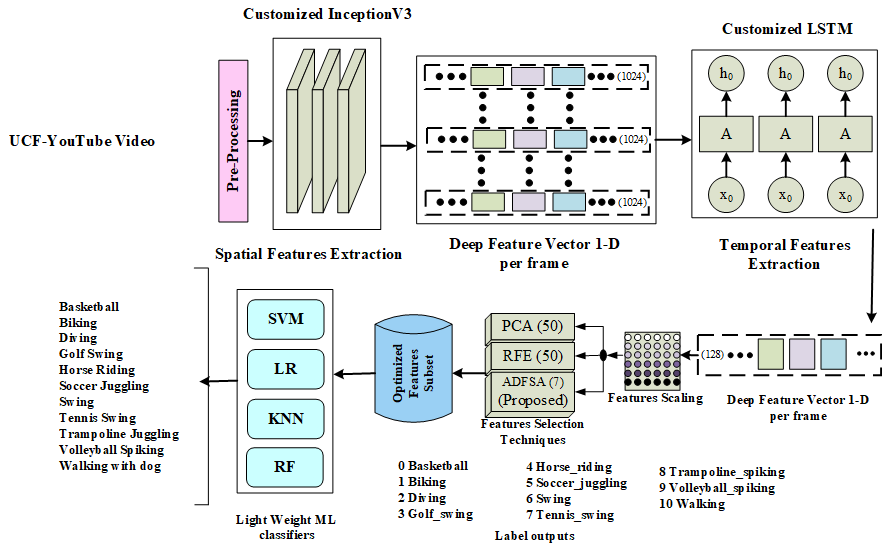} 
    \caption{\centering Overall Block Diagram of Proposed Framework.}
    \label{meth}
\end{figure}

This phased pipeline shown in Figure~\ref{meth}, ensures a balance between accuracy, compactness, and computational efficiency, supporting real-time deployment on edge devices. The framework is further benchmarked against state-of-the-art methods to validate its effectiveness.

\subsection{Data Preprocessing}
\subsubsection{Frames Extraction from Video Clip}
As deep CNN operate on image data, it is essential to extract individual frames from the video stream. To achieve this, a windowing mechanism is employed, which determines the interval at which frames are sampled and stored. This interval, referred to as the skip window size or sampling interval, is calculated using the following formula:
\begin{align}
\label{eq:W_size}
window size = \text{max}\left(\frac{\text{video\_frames\_count}}{\text{Sequence\_Length}}\right).
\end{align}

Initially, video data should undergo a frame extraction process, where individual frames are captured at a predetermined rate and resolution, in order to obtain a collection of images with uniform characteristics throughout the dataset. 

\subsubsection{Image resizing}
In this study, image preprocessing is applied to align the input dimensions of video frames with the expected input format of the modified deep convolutional architecture. Each frame is resized to $224 \times 224$ pixels with 3 color channels (RGB), ensuring compatibility with the customized Inception-based feature extractor. Alongside resizing, pixel intensities are normalized to a fixed scale, typically in the range $[0, 1]$, or standardized using the appropriate preprocessing function. This step harmonizes the input distribution with that of the network’s training conditions, promoting stable convergence and effective spatial feature extraction from each frame in the sequence.

\subsubsection{Data Augmentation}
To enhance the diversity and generalization of the model, data augmentation techniques are applied during the preprocessing phase. These augmentations simulate real-world variations and prevent overfitting by introducing randomness into the training data \cite{b29}. Common transformations include random horizontal flipping, slight rotations, zooming, and brightness adjustments, all applied within controlled bounds to preserve the semantic content of human activities. These augmentations are executed on a per-frame basis before the frames are fed into the feature extractor. By increasing variability in spatial patterns without altering temporal coherence, data augmentation enhances the robustness and reliability of temporal feature learning in downstream AA-LSTM-based modeling.

\subsection{Feature Extraction via Deep Hybrid Convolutional Neural Network}
This paper proposes a deep hybrid architecture that combines a customized InceptionV3 model with an AA-LSTM network for effective human activity recognition. The inception-V3 model is employed to extract rich spatial features from individual video frames, while the AA-LSTM network captures the temporal dependencies across frame sequences. This integration enables the model to learn both appearance-based and motion-based patterns, making it well-suited for complex video-based activity classification tasks.

\subsubsection{Spatial feature extraction via Customized Inception-V3 Network}

The proposed framework utilizes a customized inception-V3 architecture, specifically tailored to enhance spatial feature extraction for human activity recognition tasks. Modifications to the standard Inception modules include the strategic integration of pooling operations within individual branches to improve feature diversity and hierarchical abstraction. In the first Inception block, a Max-pooling (3×3) operation is added alongside Convolutional branches to capture dominant spatial activations early in the network. The second block is enhanced with both Average-pooling (3×3) after three Convolutional layers and Max-pooling (3×3) after two Convolutional layers, promoting multi-scale feature fusion. The third and fourth blocks continue this pattern with the pooling operations placed at various depths within the branches to extract increasingly abstract representations. The network concatenates all the blocks and finally passes through an 8×8 Final Global Average Pooling layer, reducing the feature map to a compact 1×1×1024 tensor. This refined design facilitates the generation of a robust 1024-dimensional feature vector that encodes discriminative spatial information crucial for downstream temporal modeling.

\begin{eqnarray}
y_{i,j} &=& \sum_{u=1}^{r} \sum_{v=1}^{c} z_{i+u-1,j+v-1} \cdot k_{u,v} \label{eq:conv}\\
y^{\text{avg}}_{i,j} &=& \frac{1}{t^2} \sum_{u=1}^{t} \sum_{v=1}^{t} z_{i+u-1,j+v-1} \label{eq:avgpool} \\
y^{\text{max}}_{i,j} &=& \max_{u=1,\ldots,t;\, v=1,\ldots,t} z_{i+u-1,j+v-1} \label{eq:maxpool}\\
Y_{i,j,:} &=& \big[ y^{(1)}_{i,j,:},\ y^{(2)}_{i,j,:},\ \ldots,\ y^{(N)}_{i,j,:} \big] \quad \text{with } D = \sum_{n=1}^N d_n \label{eq:concat} \\
h_k &=& f\left( \sum_{p} v_p W^{(1)}_{p,k} + b^{(1)}_k \right) \label{eq:fc1} \\
v_p &=& \mathrm{Flatten}(Y) \label{eq:flatten}\\ 
f_n &=& f\left( \sum_{k} h_k W^{(2)}_{k,n} + b^{(2)}_n \right) \label{eq:fc2} \\ 
\mathbf{f} &=& \big[ f_1, f_2, \ldots, f_{1024} \big]^\top \label{eq:final_feat}
\end{eqnarray}

The customized inception-V3 architecture processes spatial features through a series of convolution and pooling operations. Initially, spatial features are extracted using a standard convolution operation, as defined in Equation~\ref{eq:conv}, where the input feature map \( z \) is convolved with a kernel \( k \) of size \( r \times c \) to produce the output \( y_{i,j} \). To reduce the spatial dimensions while preserving contextual information, average pooling is applied (Equation~\ref{eq:avgpool}) by computing the mean value over a \( t \times t \) local region. Average pooling further makes stronger activations by smoothing feature maps to reduce noise from background or lighting variations, preventing overfitting by reducing sensitivity to irrelevant details, and stabilizing temporal consistency across frames, thereby retaining subtle movement cues such as those in \textquotedblleft walking.\textquotedblright

In parallel, max pooling is performed (Equation~\ref{eq:maxpool}) to capture the most salient features within the same region. It learns region-specific variations and their characteristic boundaries \cite{b30, b31}. Max-pooling selects the strongest activation functions within the pooling region, emphasizing dominant, discriminative features by capturing key motion cues such as joint positions or sharp limb movements. It increases translation invariance to small spatial shifts and reduces irrelevant background by discarding weaker activations \cite{b32}. Additionally, it speeds up computation by shrinking feature maps without losing essential details, ensuring that critical actions like the leg motion in a \textquotedblleft soccer kick\textquotedblright\ remain prominent. The outputs from multiple parallel branches (such as different filter sizes or pooling strategies) are then concatenated depth-wise to form a unified feature map \( Y \), as shown in Equation~\ref{eq:concat}, where the total number of channels \( D \) is the sum of the individual branch depths. This concatenated feature map is flattened into a 1D vector \( v \) (Equation~\ref{eq:flatten}) to prepare it for fully connected layers. The first fully connected layer transforms \( v \) into a higher-level abstract representation \( h_k \) using learned weights and a non-linear activation function \( f \), as illustrated in Equation~\ref{eq:fc1}. This is followed by a second dense layer (Equation~\ref{eq:fc2}), which refines these features into final activations \( f_n \). The complete 1024-dimensional feature vector \( \mathbf{f} \), shown in Equation~\ref{eq:final_feat}, serves as the final spatial representation extracted from the image and is subsequently passed to the temporal modeling module, such as an AA-LSTM network.

\subsubsection{Spatio-Temporal Representation via Attention-Augmented LSTM}

Following the extraction of spatial features from each frame using the customized InceptionV3 model, sequences of 20 consecutive 1024-dimensional vectors, representing one video segment, are used to train an AA-LSTM network. The AA-LSTM is designed with 128 hidden units and is trained to model temporal dependencies across the sequential spatial features, learning dynamic transitions and contextual patterns associated with various human activities. After training, the AA-LSTM is detached from its classification objective and repurposed as a temporal feature extractor. For each input sequence of shape $(20, 1024)$, the trained AA-LSTM processes the temporal structure and outputs a 128-dimensional feature vector corresponding to its final hidden state. This vector serves as a compact spatial–temporal representation, enriched with both appearance-based and motion-based information, and is used in subsequent stages such as feature fusion or classification.
\begin{eqnarray}
\mathbf{F} &=& \left[ \mathbf{F}_1, \mathbf{F}_2, \dots, \mathbf{F}_T \right], \quad \mathbf{F}_t \in \mathbb{R}^{1024} \label{eq:input_sequence} \\[4pt]
\left[ \mathbf{i}_t, \mathbf{f}_t, \mathbf{o}_t, \tilde{\mathbf{c}}_t \right] &=& \sigma \left( \mathbf{W} \cdot \left[ \mathbf{h}_{t-1}, \mathbf{F}_t \right] + \mathbf{b} \right) \label{eq:gates} \\[4pt]
\mathbf{c}_t &=& \mathbf{f}_t \odot \mathbf{c}_{t-1} + \mathbf{i}_t \odot \tilde{\mathbf{c}}_t \label{eq:cell_state} \\
\mathbf{h}_t &=& \mathbf{o}_t \odot \mathit{tanh}(\mathbf{c}_t) \label{eq:hidden_state} \\[4pt]
e_t &=& \mathbf{v}^\top \mathit{tanh}\left( \mathbf{W}_a \cdot \mathbf{h}_t + \mathbf{b}_a \right) \label{eq:score}\\ [4pt]
\alpha_t &=& \frac{\exp(e_t)}{\sum_{k=1}^{T} \exp(e_k)} \label{eq:alpha}\\ [4pt]
\mathbf{T} &=& \sum_{t=1}^{T} \alpha_t \cdot \mathbf{h}_t \in \mathbb{R}^{128} \label{eq:temporal_embedding}
\end{eqnarray}

To effectively model both spatial and temporal dependencies within a video, frame-level spatial features $\mathbf{F}_t \in \mathbb{R}^{1024}$ are first extracted for each of the $T$ frames, forming a sequential input $\mathbf{F} = [\mathbf{F}_1, \mathbf{F}_2, \dots, \mathbf{F}_T]$ as defined in Equation~\ref{eq:input_sequence}. These features are then fed into a customized AA-LSTM network, which learns temporal transitions across time steps. At each time step $t$, the AA-LSTM computes the input gate $\mathbf{i}_t$, forget gate $\mathbf{f}_t$, output gate $\mathbf{o}_t$, and the candidate cell state $\tilde{\mathbf{c}}t$ based on the previous hidden state $\mathbf{h}{t-1}$ and the current input $\mathbf{F}_t$, as shown in Equation~\ref{eq:gates}. The cell state $\mathbf{c}_t$ is updated using Equation~\ref{eq:cell_state}, and the corresponding hidden state $\mathbf{h}_t$ is computed according to Equation~\ref{eq:hidden_state}.

To enable the model to focus selectively on important time steps, an attention mechanism is integrated. Alignment scores $e_t$ are calculated using a feedforward layer with a learnable attention weight vector $\mathbf{v}$, as described in Equation~\ref{eq:score}. These scores are then normalized using the softmax function to obtain attention weights $\alpha_t$ (Equation~\ref{eq:alpha}), which reflect the importance of each hidden state $\mathbf{h}_t$ in the overall sequence context. The final temporal representation $\mathbf{T} \in \mathbb{R}^{128}$ is obtained as a weighted sum of the hidden states (Equation~\ref{eq:temporal_embedding}), serving as a compact, attention-enhanced encoding of the video sequence.

\begin{figure}[!htbp]
    \centering
    \includegraphics[width=1\textwidth]{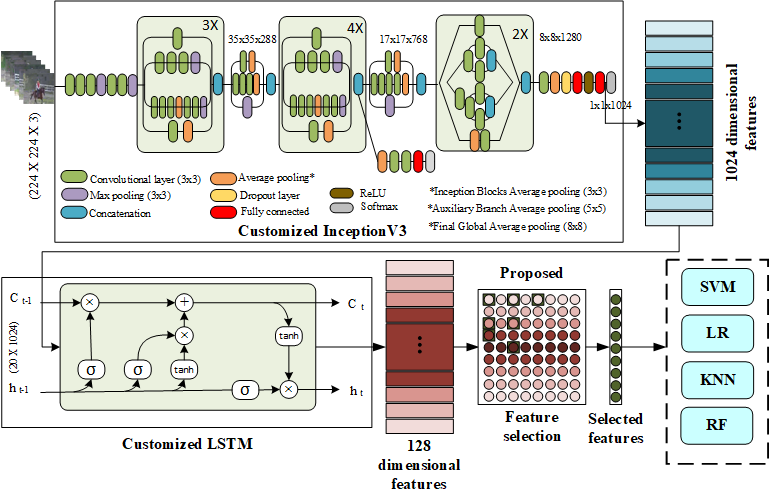} 
    \caption{\centering Detailed overview of Proposed Framework}
    \label{spatem}
\end{figure}

\subsection{Features Selection}
FS plays a critical role in the development of efficient and robust machine learning models, particularly in high-dimensional datasets such as those used in video-based HAR. By identifying and retaining only the most relevant and informative features, FS helps reduce model complexity, minimize overfitting, improve generalization, and significantly lower inference time. It also enhances the interpretability of the model by focusing on features that contribute most to the classification task. The FS process typically involves evaluating the relevance, redundancy, and contribution of each feature using statistical measures, model-based techniques, or optimization algorithms. In this study, a systematic FS approach was applied, which began with extracting a comprehensive set of spatial and temporal features. These features were then evaluated using methods such as PCA \cite{b33}, RFE \cite{b34}, and a proposed dynamic composite fitness function to identify the most discriminative subset. The selected features were subsequently used to train lightweight classifiers, ensuring both high accuracy and computational efficiency.

\subsubsection{Dynamic Fitness Sharing for Feature Selection}
During the process of FS, the main population is divided into sub-populations to prevent an excessive increase in individuals within a specific high point. This strategy results in individuals gradually shifting from one peak to another peak, as their numbers rise. The method achieves this by lowering the fitness score of similar individuals as their population grows. It is assumed that similar individuals belong to the same group and occupy the same niche. A similarity measure and a threshold (niche radius) are established in the search space to determine when individuals are considered similar and part of the same niche. Studies have shown that with a sufficiently large population size and appropriately set niche radius, this approach can create as many species as there are peaks in the fitness landscape, effectively covering all niches \cite{b35}.

To counteract the selection pressure issue in Fitness Sharing methods, the fitness scores of individuals within a particular peak or niche are modified. This adjustment is based on the density of similar individuals in the vicinity, operating under the assumption that individuals with similar traits belong to the same niche. The similarity between two individuals is determined using a metric, which can be applied in either the genotypic or phenotypic space. A threshold value is then used to define the maximum permissible distance for individuals to be considered similar and part of the same niche \cite{b35}. The modified DFS is discussed in the coming section in detail.

\subsubsection{Proposed Novel composite Adaptive Dynamic Fitness Sharing with Attention Mechanism (ADFSA)}
A dynamic composite fitness function is proposed for feature set optimization in HAR. The proposed technique integrates multiple criteria, classification accuracy, feature redundancy, uniqueness, and complexity into a unified optimization framework using a dynamically adaptive weighting strategy. This dynamic composite fitness function is designed to guide the give GA-based search process to identify an optimal subset of features that are both discriminative and compact.

\begin{figure}[!htbp]
    \centering
    \includegraphics[width=0.8\textwidth]{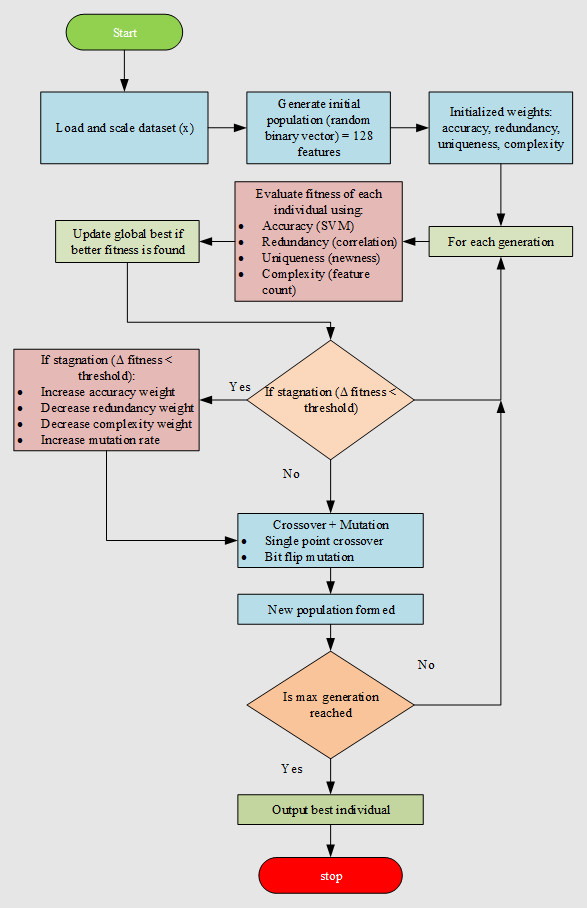} 
    \caption{\centering Flow of GA incorporated Proposed dynamic composite ADFSA}
    \label{Flow_GA}
\end{figure}

The ADFSA technique involves three stages:
\begin{enumerate}
\item Adaptive niching Based on Feature Redundancy: Creating 'niches' dynamically will introduce redundancy in characteristics that cannot be eliminated only by considering standard distance metrics such as the Hammer distance. In the final selection, penalizing the feature subsets that have too many similar features leads to model performance degradation and overfitting. A feature subset having high redundancy can be computed by correlation and mutual information between features, and applying a redundancy penalty.
\item Uniqueness-Based Exploration: Those prioritized features that were not explored in the previous generations are assigned a uniqueness score. The idea is to penalize common subsets and reward unique ones that have not been sampled frequently. Keep track of the selected features across the generations and reward with the subset features that deviate from historical selections. 
\item Hybrid Objective Optimization: In DFS, typically focuses on optimizing accuracy (or another performance metric), introducing uniqueness will combine accuracy with model interoperability or computation cost as an addition. For example, rewarding features that lead to simpler models (fewer features, lower computation costs) while preserving accuracy leads to a hybrid fitness function. The hybrid fitness function combines accuracy, redundancy, and model complexity. The overall flow of the proposed FS is given in Figure~\ref {Flow_GA} 
\end{enumerate}

\subsubsection{Mathematical Interpretation of our Proposed ADFSA Technique}
The proposed technique combines multiple factors, accuracy, redundancy, uniqueness-boost, and complexity into a dynamic FS framework. The following is the mathematical representation of the ADFSA scheme incorporated in the proposed framework.

\begin{enumerate}
\item Fitness Function: The fitness function evaluates the feature subset (as a binary vector representation) and aims for high accuracy optimization by penalizing complexity and redundancy. Mathematically, it can be represented as:

\begin{eqnarray}
\label{fit_ftn}
\mathcal{F}(S) = \alpha(g) \cdot f_{acc}(S) - \beta(g) \cdot f_{red}(S)  + \gamma(g) \cdot f_{uni}(S) - \delta(g) \cdot f_{comp}(S)
\end{eqnarray}

where $\alpha(g)$, $\beta(g)$, $\gamma(g)$, and $\delta(g)$ denote the generation-dependent weights assigned to the corresponding components of the fitness function, these weights dynamically control the influence of classification accuracy, redundancy penalty, uniqueness-boost, and complexity penalty, respectively. Typically, the weights satisfy the constraint $\alpha(g) + \beta(g) + \gamma(g) + \delta(g) = 1$, and are adjusted empirically based on the optimization stage. The binary vector $S$ represents the selected feature subset, where a value of 1 indicates a selected feature and 0 indicates otherwise. The terms $f_{\text{acc}}$, $f_{\text{red}}$, $f_{\text{uni}}$, and $f_{\text{comp}}$ correspond to the classification accuracy, redundancy measure, uniqueness score, and complexity measure, respectively, as defined in Equations~(2) through (5).

\item \textbf{Accuracy Term}: This term measures the ability of the performance of a selected feature subset in terms of classification, often calculated by cross-validation on a training set as follows:

\begin{eqnarray}
    f_{acc}(S) = \frac{1}{k} \sum_{i=1}^{k} {Accuracy}(S_i)
\end{eqnarray}

where $S_i$ represents the selected subset of features at the i$-th$ cross-validation fold and $K$ is the cross-validation fold number.

\item \textbf{Redundancy Penalty}: 
The redundancy score $f_{\text{red}}(S)$ is used to quantify the degree of correlation among the selected features in the subset $S$. It is computed based on the Pearson correlation coefficient $r_{ij}$ between the $i$-th and $j$-th features. If the absolute correlation $|r_{ij}|$ exceeds a predefined threshold $t$ (e.g., 0.7), the corresponding feature pair is considered highly redundant. This is captured using an indicator function $\mathbb{I}(|r_{ij}| > t)$, which returns 1 when the condition is true and 0 otherwise. The final redundancy score is normalized over all unique feature pairs and is given by:
\begin{eqnarray}
    f_{red}(S) = \frac{1}{n(n-1)} \sum_{i=1}^{n-1} \sum_{j=i+1}^{n} \mathbb{I}(|r_{ij}| > t)
\end{eqnarray}

A higher value of $f_{\text{red}}(S)$ indicates more redundancy among selected features, which is penalized in the fitness evaluation to encourage diversity and minimize overlap among feature contributions. 

\item \textbf{Uniqueness-Boost:} To encourage exploration and maintain population diversity, a uniqueness reward is introduced. This reward is granted to feature subsets that have not been encountered in prior generations. It is computed as:
\begin{equation}
    f_{uni}(S) = \mathbb{I}\{ S \notin \mathcal{H} \}
\end{equation}

where $\mathcal{H}$ denotes the set of previously evaluated feature subsets (i.e., the search history), and $\mathbb{I}\{ S \notin \mathcal{H} \}$ is an indicator function that returns 1 if the subset $S$ is novel (not present in history), and 0 otherwise. This mechanism rewards diversity by promoting unseen solutions, thus helping to avoid premature convergence of the algorithm.

\item \textbf{Complexity Penalty:} This component discourages the selection of excessively large feature subsets, promoting model simplicity and reducing overfitting. It is calculated as:
\begin{eqnarray}
    f_{comp}(S) = \frac{|S|}{n}
\end{eqnarray}

where $|S|$ represents the number of selected features in subset $S$ (i.e., the number of 1's in the binary selection vector), and $n$ denotes the total number of available features. A higher value of $f_{\text{comp}}$ indicates a more complex subset, which is penalized in the fitness function to favor parsimonious models.
\end{enumerate}

To improve convergence and enhance search efficiency, the weights of the dynamic composite fitness function are dynamically adapted throughout the evolutionary process. The goal is to favor exploration (diversity and redundancy control) in early generations and gradually shift focus toward exploitation (accuracy and compactness) in later generations.

Let $G$ be the total number of generations and $g \in [1, G]$ be the current generation.

Define the normalized progress as:
\begin{eqnarray}
    p = \frac{g}{G}
    \label{eq:progress}
\end{eqnarray}

Then the weights are updated as follows:
\begin{eqnarray}
    \alpha(g) & = & 0.4 + 0.6 \cdot p \quad \text{(from 0.4 to 1.0)}\label{eq:alpha}\\
    \beta(g) & = & 0.3 - 0.2 \cdot p \quad \text{(from 0.3 to 0.1)}\label{eq:beta}\\
    \gamma(g) & = & 0.3 - 0.2 \cdot p \quad \text{(from 0.3 to 0.1)} \label{eq:gamma} \\
    \delta(g) & = & 0.2 + 0.1 \cdot p \quad \text{(from 0.2 to 0.3)} \label{eq:delta}
\end{eqnarray}

In the proposed fitness function, the dynamic behavior of the evolutionary algorithm is controlled through four generation-dependent weights: $\alpha(g)$, $\beta(g)$, $\gamma(g)$, and $\delta(g)$. These weights adapt over time according to the normalized generation progress $p$, defined in Equation~\eqref{eq:progress}, where $g$ is the current generation and $G$ is the total number of generations.

The weight $\alpha(g)$ in Equation~\eqref{eq:alpha} increases linearly from 0.4 to 1.0 as $p$ grows, placing greater emphasis on the accuracy component of the fitness function in later generations. This encourages exploitation once the algorithm has sufficiently explored the search space. Conversely, both $\beta(g)$ and $\gamma(g)$, as shown in Equations~\eqref{eq:beta} and~\eqref{eq:gamma}, decrease linearly from 0.3 to 0.1, reducing the influence of redundancy and novelty as evolution progresses. This shift reflects the reduced need for diversity and exploration in later stages. Lastly, $\delta(g)$ in Equation~\eqref{eq:delta} increases slightly from 0.2 to 0.3 to mildly penalize overly complex feature subsets, promoting compact representations toward the end of the search.

This adaptive scheme ensures a smooth transition from exploration to exploitation during FS, leading to more optimal and generalizable feature subsets.

\begin{algorithm}[!h]
\footnotesize
\caption{Evolutionary FS with Dynamic Adjustment}
\begin{algorithmic}
\STATE Set $previous\_best\_fitness \leftarrow 0$
\STATE Set $generation \leftarrow 0$
\WHILE{$generation < n\_generations$}
    \IF{$generation > 5$ AND $max(fitness\_scores) - previous\_best\_fitness < 0.001$}
        \STATE $accuracy\_weight \leftarrow \min(accuracy\_weight + 0.05, 1.0)$
        \STATE $redundancy\_weight \leftarrow \max(redundancy\_weight - 0.05, 0.1)$
        \STATE $mutation\_rate \leftarrow \min(mutation\_rate + 0.01, 0.2)$
    \ENDIF
    \STATE Initialize $fitness\_scores \leftarrow []$
    \FOR{each $individual$ in $population$}
        \STATE \textbf{$fitness \leftarrow evaluate\_individual(individual, x,y,w)$} \COMMENT{Fitness function is applied here.}
        \STATE Append $fitness$ to $fitness\_scores$
    \ENDFOR
    \STATE \textbf{Sort $population$ based on $fitness\_scores$} \COMMENT{Fitness scores determine sorting.}
    \STATE Select top half of $population$
    \STATE Initialize $new\_population \leftarrow []$
    \WHILE{length($new\_population$) < $population\_size$}
        \STATE Randomly select two parents: $parent1$, $parent2$ from $population$
        \STATE $crossover\_point \leftarrow$ random integer between $1$ and $n\_features - 1$
        \STATE $child1 \leftarrow$ combine $parent1$ and $parent2$ at $crossover\_point$
        \STATE $child2 \leftarrow$ combine $parent2$ and $parent1$ at $crossover\_point$
        \FOR{each $child$ in [$child1$, $child2$]}
            \IF{random probability < $mutation\_rate$}
                \STATE $mutation\_point \leftarrow$ random integer between $0$ and $n\_features - 1$
                \STATE Flip the value of $child[mutation\_point]$
            \ENDIF
        \ENDFOR
        \STATE Append $child1$ and $child2$ to $new\_population$
    \ENDWHILE
    \STATE Set $population \leftarrow new\_population$
    \STATE \textbf{Set $previous\_best\_fitness \leftarrow max(fitness\_scores)$} \COMMENT{Updating based on fitness values.}
    \STATE Print "Generation ", $generation$, " complete, top fitness: ", $previous\_best\_fitness$
    \STATE Increment $generation \leftarrow generation + 1$
\ENDWHILE
\end{algorithmic}
\end{algorithm}

\subsubsection{Key Advantages of the Proposed Framework}
The proposed framework introduces a comprehensive and adaptive FS technique that integrates four fundamental aspects, \textit{accuracy}, \textit{redundancy}, \textit{uniqueness}, and \textit{complexity} into a unified fitness function. This multi-objective formulation ensures a balanced trade-off between classification performance and model simplicity. One of the key innovations lies in the dynamic adjustment of the weights associated with each fitness component during the evolutionary process. This adaptive mechanism fosters exploration in the initial generations and shifts towards exploitation in later stages, which enhances convergence and helps avoid premature stagnation in local optima. Additionally, the method incorporates a redundancy penalty based on mutual information, effectively minimizing the selection of highly correlated features and ensuring that only the most informative attributes are retained. To further improve population diversity, a uniqueness term is introduced that rewards diverse feature subsets across generations, increasing the likelihood of identifying globally optimal solutions. The inclusion of a complexity component drives the selection of smaller, more efficient feature sets, thereby reducing the computational cost and facilitating faster model inference, an essential requirement for real-time and embedded HAR systems. Empirical evaluations demonstrate that the proposed technique outperforms traditional FS techniques such as PCA and RFE in terms of classification accuracy, even with fewer selected features. Moreover, the approach is classifier-agnostic and compatible with a variety of learning algorithms, including SVM, KNN, Random Forest (RF), and Logistic Regression (LogReg). Its scalability makes it particularly suitable for handling high-dimensional feature vectors derived from deep learning models, addressing the limitations of conventional methods related to scalability and interpretability. Overall, the proposed composite, dynamically adaptive, and diversity-aware technique presents a novel and effective solution for FS in HAR and other pattern recognition domains.

In our proposed technique, redundancy and uniqueness-boost mechanisms ensure that the algorithm does not converge too early on a suboptimal feature subset. Furthermore, the inclusion of model complexity into the fitness function will produce simpler models, which are more computationally efficient and easier to interpret. Incorporating flexibility across different types of FS tasks, and penalty terms having adaptive natures, allows fine-tuning based on the specific problem or dataset. 

\subsection{Classification}

To assess the effectiveness of the proposed FS technique, we perform a comparative evaluation against two widely used techniques: PCA and RFE. PCA is a linear dimensionality reduction approach that transforms the feature space into orthogonal components capturing the maximum variance. RFE, on the other hand, iteratively removes the least important features based on model weights, gradually selecting the most significant subset.

Our proposed technique employs a custom-designed fitness function that adaptively balances classification accuracy $f_{acc}$, feature redundancy $f_{red}$, uniqueness $f_{uni}$, and complexity $f_{comp}$. The dynamic weighting mechanism evolves over generations, initially favoring diversity and redundancy control, and later shifting focus toward accuracy and compactness.

To evaluate classification performance and testing efficiency, the optimized features obtained from each selection method are fed into a set of lightweight machine learning classifiers, including SVM, RF, KNN, and LogReg. SVM is chosen for its strong generalization capability in high-dimensional spaces. RF offers robustness to overfitting and effectively handles noisy or complex data. k-NN serves as a non-parametric baseline suitable for low-dimensional feature spaces, while LR provides a fast, interpretable linear model as a benchmark.

Let the optimized feature vector selected by the GA be denoted as:
\begin{eqnarray}
\mathbf{f}^* &=& [f^*_1, f^*_2, \ldots, f^*_d]^\top \in \mathbb{R}^d \\
\hat{Y}_\text{{SVM}} &=& \mathrm{sign}(\mathbf{w}^\top \mathbf{f}^* + b)
\label{eq:svm}\\
\hat{Y}_{\text{RF}} &=& \mathrm{mode}\big( \{ h_t(\mathbf{f}^*) \}_{t=1}^T \big)
\label{eq:rf}\\
\hat{Y}_{\text{kNN}} &=& \mathrm{mode}\big( \{ y_i : \mathbf{f}_i \in \mathcal{N}_k(\mathbf{f}^*) \} \big)\label{eq:knn} \\
\hat{Y}_{\text{LR}} &=& \begin{cases}
1, & \text{if } \sigma(\mathbf{w}^\top \mathbf{f}^* + b) \geq 0.5 \\
0, & \text{otherwise}
\end{cases}
\label{eq:lr}
\end{eqnarray}
The optimized feature vector $\mathbf{f}^*$, generated through the proposed GA-based FS technique, is evaluated using multiple lightweight machine learning classifiers to assess its classification capability and generalizability. The SVM, as formulated in Equation~\ref{eq:svm}, determines the class label by computing the sign of a weighted linear combination of features and bias, effectively creating a maximum-margin decision boundary. The RF classifier, shown in Equation~\ref{eq:rf}, aggregates predictions from an ensemble of $T$ decision trees and outputs the majority-voted label, thereby reducing variance and enhancing robustness. The k-Nearest Neighbors (k-NN) approach, expressed in Equation~\ref{eq:knn}, relies on the most frequent class among the $k$ closest samples in the feature space, making it highly adaptive to local patterns in the data. LogReg, represented in Equation~\ref{eq:lr}, calculates the sigmoid-activated probability and assigns the class label based on a threshold decision rule. Together, these classifiers provide a diverse set of perspectives to validate the quality of the selected features. To further benchmark the effectiveness of the proposed FS technique, classical techniques like PCA and RFE are also applied under identical evaluation settings. All models are evaluated in terms of both classification accuracy and inference time to validate the practical efficiency of the selected features.

\section{Experimental Setup}

\begin{figure}[!htbp]
    \centering
    \includegraphics[width=1.0\textwidth]{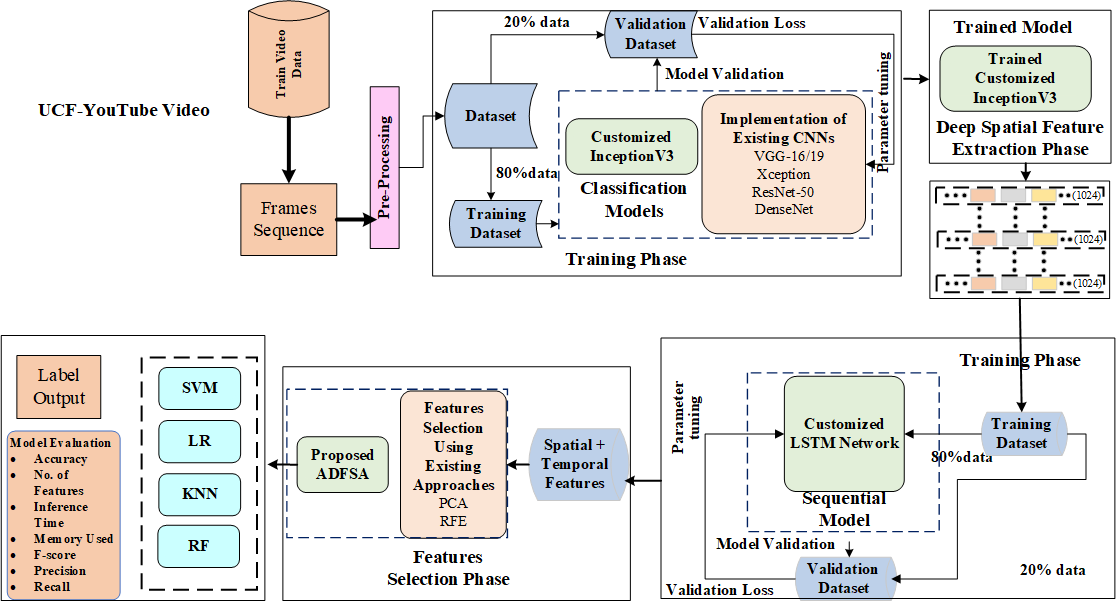} 
    \caption{\centering Experimental setup of Proposed Framework}
    \label{framework}
\end{figure}

The experimental setup of this research is structured into four main components as given in the Figure~\ref{framework}. First, the dataset is introduced, highlighting its composition and class distribution. Second, a two-phase framework is applied, where spatio-temporal features are extracted using a customized deep learning architecture and then refined through the proposed feature selection technique. Third, the hardware and software environment is outlined to specify the implementation platform and computational resources used. Together, these components provide a clear foundation for evaluating the effectiveness and efficiency of the proposed framework.

\subsection{Dataset}
The YouTube Action Dataset \cite{b11} is a widely used benchmark in human activity recognition, consisting of diverse videos collected from YouTube that capture real-world human actions in varied contexts and environments. It covers multiple action categories such as walking, running, biking, and hand-waving, among others, making it a challenging testbed for evaluating recognition algorithms. In this research, the dataset was divided into training and testing sets using an $80/20$ split to ensure balanced evaluation of the proposed framework.

What sets this dataset apart is its variability in camera angles, background clutter, lighting conditions, and motion dynamics, which better reflect real-world scenarios compared to more controlled datasets. Each video sequence is typically labeled with a single dominant activity, and many clips that test the robustness of recognition models.

Due to its complexity and diversity, the YouTube Action Dataset has become a standard benchmark for assessing the effectiveness of deep learning architectures, spatiotemporal modeling techniques, and feature extraction frameworks in video-based activity recognition. The detailed configuration of the dataset is outlined in Table \ref{tab:ucf11_config}.
\begin{table}[!h]
\centering
\small
\caption{\centering Configuration of the UCF-YouTube (UCF11) Dataset \cite{b11}}
\begin{tabular}{|l|p{10cm}|}
\hline
\textbf{Attribute} & \textbf{Details} \\
\hline
Full Name & UCF YouTube Action Dataset (UCF11) \\
\hline
Source & YouTube videos \\
\hline
Total Action Classes & 11 \\
\hline
Action Classes & Basketball Shooting, Biking, Diving, Golf Swing, Horse Riding, Soccer Juggling, Swinging, Tennis Swing, Trampoline Jumping, Volleyball Spiking, Walking with a Dog \\
\hline
Total Video Clips & 1,600 \\
\hline
Clips per Class & Approximately 145 (varies slightly) \\
\hline
Resolution & 320×240 (typical, but varies) \\
\hline
Frame Rate & 29.97 fps (varies) \\
\hline
Video Format & .avi \\
\hline
Total Groups & 25 per class (each group has different people/backgrounds/camera motions) \\
\hline
Group-wise Split & Leave-one-group-out cross-validation commonly used \\
\hline
Duration per Clip & Approximately 4–10 seconds \\
\hline
Labels & Provided via folder structure (one folder per class) \\
\hline
Modality & RGB (no audio/depth/skeleton) \\
\hline
Tasks Supported & Action Recognition, Temporal Segmentation \\
\hline
Preprocessing Required & Frame extraction, resizing, normalization \\
\hline
Common Benchmarks & Accuracy, Precision, Recall, F1-score \\
\hline
Download URL & \url{http://crcv.ucf.edu/data/UCF_YouTube_Action.php} \\
\hline
\end{tabular}
\label{tab:ucf11_config}
\end{table}

\subsection{Hardware and Implementation}

All algorithms, including feature selection techniques and classifiers such as SVM, RF, KNN, and LogReg, were implemented using Python 3.10 within the Spyder integrated development environment (IDE). The codebase utilized well-established libraries such as \texttt{TensorFlow/Keras}, \texttt{Scikit-learn}, \texttt{OpenCV}, and \texttt{NumPy}. For conventional ML experiments, a workstation with an Intel Core i7 10th-generation processor and 16 GB of RAM was used. In contrast, training of computationally demanding deep models was performed on a high-performance HP desktop, equipped with an Intel Core i9 9th-generation processor, 64 GB of RAM, and an NVIDIA GeForce RTX 2080 Ti GPU.  

The deep learning framework consisted of a customized InceptionV3 backbone integrated with an AA-LSTM temporal model. Input frames were resized to $224 \times 224 \times 3$, which is a widely adopted dimension that provides a balance between computational efficiency and the preservation of discriminative spatial details. Each video clip was represented by 20 frames to capture sufficient temporal information while avoiding excessive redundancy. The Adam optimizer with a learning rate of $1 \times 10^{-4}$ was selected due to its robustness in handling sparse gradients and faster convergence in deep architectures. A batch size of 32 was chosen to maintain a trade-off between training stability and GPU memory constraints. Categorical cross-entropy was employed as the loss function since it is well-suited for multi-class classification problems. Finally, the models were trained for 100 epochs, providing enough iterations to achieve convergence without leading to overfitting.  

The complete hardware and hyperparameter configuration is summarized in Table~\ref{tab:Setup}. This dual-platform and carefully tuned hyperparameter setup ensured efficient execution of lightweight ML tasks as well as effective training of compute-intensive deep learning models.  

All machine learning classifiers and FS algorithms were implemented and executed using the Spyder integrated development environment (IDE).The codebase was implemented in python 3.10, utilizing \texttt{TensorFlow/Keras}, \texttt{Scikit-learn}, \texttt{openCV}, and \texttt{NumPy} libraries. The experiments related to traditional ML models, such as SVM, RF, KNN, and LogReg, were conducted on a system equipped with an Intel Core i7 10th-generation processor and 16 GB of RAM. For the training of deep CNN, including customized InceptionV3 and AA-LSTM-based models, a high-performance computing setup was used, consisting of an HP desktop powered by an Intel Core-i9 9th-generation processor, 64 GB of RAM, and an NVIDIA GeForce RTX 2080 Ti GPU, all the cofiguration is given in Table~\ref{tab:Setup}. This dual-platform configuration was essential to accommodate both lightweight ML tasks and compute-intensive deep learning workloads effectively. 

\begin{table}[!h]
\begin{center}
\caption{\centering Configuration details of the experimental setup.}
\begin{tabular}{|l|c|}
\hline
\textbf{Component} & \textbf{Details} \\ \hline
Backbone Network & Customized inceptionV3 (pretrained) \\ \hline
Temporal Model   & Customized AA-LSTM with 128 units \\ \hline
Input Size       & $224 \times 224 \times 3$ \\ \hline
Frames per Video & 20 \\ \hline
Optimizer        & Adam ($\text{lr} = 1\times10^{-4}$) \\ \hline
Batch Size       & 32 \\ \hline
Loss Function    & Categorical Cross-Entropy \\ \hline
Training Epochs  & 100 \\ \hline
Hardware         & NVIDIA RTX 2080 Ti, 32 GB RAM \\ \hline
\end{tabular}
\label{tab:Setup}
\end{center}
\end{table}

\section{Results and Discussions}
Extensive experiments were conducted on benchmark HAR datasets to evaluate the proposed dynamic composite fitness function for feature selection. Compared with PCA and RFE, the method achieved higher accuracy, selected fewer features, and improved inference speed. By adjusting the balance between accuracy, redundancy, uniqueness, and complexity, the fitness function guided the search toward compact yet effective feature subsets. In one case, only 7 out of 128 features were selected while still reaching a test accuracy of 99.58\%, along with reduced memory use and faster execution. These results show that the method can provide accurate and efficient recognition, even in real-time or resource-limited settings.

The detailed results are presented in the following subsections. First, two-phase features extraction and selection framework is dicussed, and 2D feature space visualizations are shown to highlight the separability of the optimized features. Next, evaluation metrics and implementation settings are described. This is followed by a performance summary along with memory and training time analysis. Additional measures such as F1-score, precision, and recall are then reported to provide a broader view of classification performance. An ablation study is included to examine the contribution of each component of the fitness function. Finally, the proposed technique is compared with state-of-the-art approaches to demonstrate its overall effectiveness.

\subsection{Two-Phase Feature Extraction and Selection Framework}
Overall, the research was implemented into two different phases. In the first phase, spatio-temporal features were extracted using a customized Inception-V3 and AA-LSTM network, and in the second phase, the extracted features were passed through the proposed ADFSA FS technique to obtain compact and robust features without affecting the performance of the underlying classifier. 

\subsubsection{Spatial Feature Extraction Phase}

\begin{figure}[!htbp]
    \centering
    \begin{subfigure}{0.48\textwidth}
        \includegraphics[width=\linewidth]{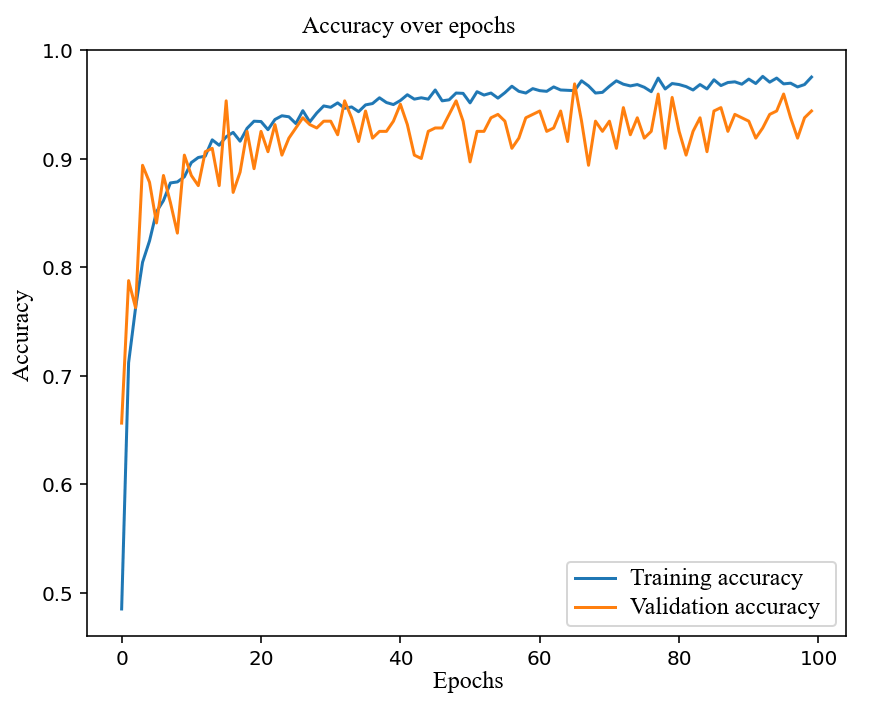}
        \caption{\centering Model Accuracy.}
        \label{fig:accuracy}
    \end{subfigure}
    \hspace{0.5em} % Reduce space between the two subfigures
    \begin{subfigure}{0.48\textwidth}
        \includegraphics[width=\linewidth]{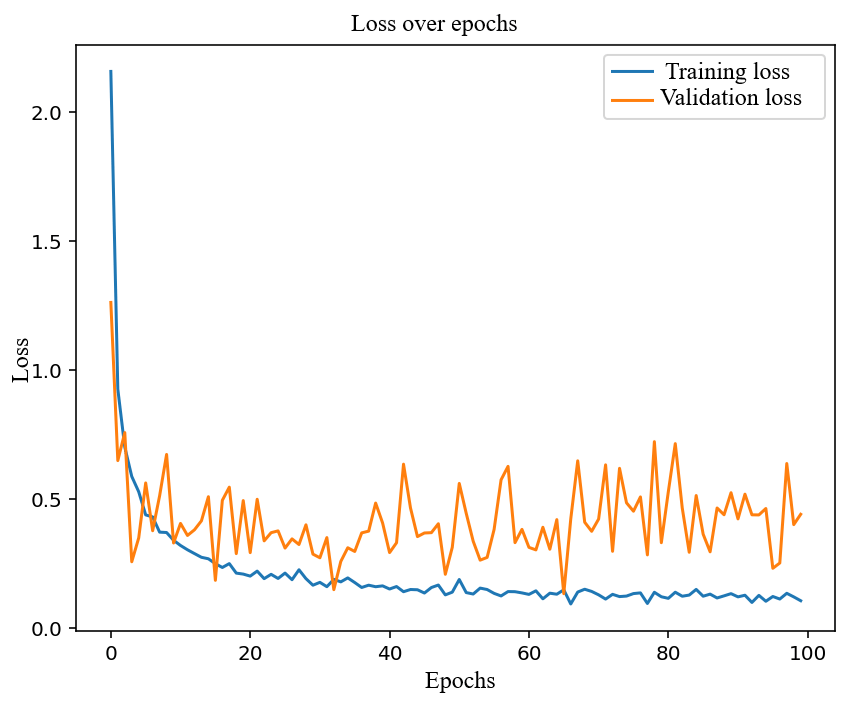}
        \caption{\centering Model Loss.}
        \label{fig:loss}
    \end{subfigure}
    \caption{\centering InceptionV3 Model Accuracy and Loss during Training.}
    \label{fig:feature_comparison}
\end{figure}

This work employs a deep CNN architecture to extract discriminative features from video frames. Extensive experiments were carried out by training several deep CNN models, as summarized in Table~\ref{tab:D-CNN}. Among the evaluated architectures, InceptionV3 consistently outperformed the others, demonstrating good recognition accuracy. Trainng of selected networks and the overall proposed framework is presented in Figure \ref{framework}. Consequently, it was selected and customized for the HAR task, capturing diverse and robust features region and edge, leading to enhanced performance. The training and loss curves of the optimized model are illustrated in Figures~\ref{fig:accuracy} and~\ref{fig:loss}, respectively.

In the feature extraction stage, each input video is decomposed into a sequence of frames at a fixed sampling rate to capture meaningful visual cues. These frames are resized and normalized before being passed through the customized InceptionV3 network, which is employed as a deep feature extractor. Specifically, the frames are processed up to the final global average pooling layer, producing a 1024-dimensional feature vector per frame. These features encode high-level spatial representations of the visual content, capturing semantic attributes such as shapes, textures, and object structures. The extracted frame-wise features are then stored sequentially to preserve their temporal order for subsequent processing stages.

\subsubsection{Temporal Feature Learning using AA-LSTM}
To capture the temporal dynamics present across consecutive video frames, an AA-LSTM network is employed. Features of 20 frames extracted from customized Inception-V3 in a sequential manner are fed into the AA-LSTM, learning long-range dependencies between consecutive individual frames, retaining historical context over time. In this setup, the AA-LSTM network consists of 128 memory units and is trained using the frame-wise feature vectors in sequence. This enables the model to understand motion patterns, temporal transitions, and dependencies between frames that are critical for activity recognition. After training, the AA-LSTM outputs a feature representation that integrates both spatial information (from InceptionV3) and temporal dependencies (learned via AA-LSTM). These fused spatio-temporal features are then used for subsequent feature optimization and classification tasks.

\begin{table}[!h]
\centering
\small
\caption{\centering Different selected Deep CNN}
\begin{tabular}{|l|c|}
\hline
\textbf{Methods} & \textbf{Accuracy} \\
\hline
Xception & 94.42\% \\
VGG-16 & 95.35\% \\
VGG-19 & 95.63\% \\
Resnet-50 & 88.34\% \\
ResNet50-v2 & 98.73\% \\
Resnet-152 & 81.24\% \\
ResNet-101 & 78.33\% \\
DenseNet-121 & 96.89\% \\
DenseNet-169 & 96.90\% \\
NasNet-Large & 96.86\% \\
Inception-V3 & 97.07\% \\
\hline
\textbf{Customized Inception-V3} & \textbf{98.15\%} \\
\textbf{Customized Inception-V3 + AA-LSTM} & \textbf{99.32\%} \\
\hline
\end{tabular}
\label{tab:D-CNN}
\end{table}

\subsubsection{Features Selection via ADFSA}
To enhance the quality of the extracted features, an optimization step was carried out using the proposed ADFSA algorithm. This method leverages a GA guided by a dynamic composite fitness function that integrates four complementary evaluation criteria. First, \textit{classification accuracy} is assessed using stratified 5-fold cross-validation with a linear SVM, ensuring robust performance estimation. Second, a \textit{redundancy penalty} based on mutual information is incorporated to discourage the selection of highly correlated features, promoting a more informative subset. Third, a \textit{uniqueness component} is introduced to enhance population diversity by penalizing the recurrence of identical feature subsets across generations. Finally, a \textit{complexity term} penalizes larger subsets, thereby encouraging compact and computationally efficient FSs. Notably, the relative weights of these components are adaptively updated throughout the evolutionary process, guided by the normalized progression of the population's fitness. This dynamic adjustment ensures a balance between exploration and exploitation, ultimately leading to the selection of high-quality, low-redundancy feature sets well-suited for classification tasks. To assess the effectiveness of the proposed technique, two widely used FS techniques were implemented, PCA \cite{b25} and RFE \cite{b26}. The effectiveness of various selection methods was assessed using a range of classification algorithms, comprising SVM, RF, KNN, and LogReg. 

\subsection{2D Features Space Visualization}

\begin{figure}[!htbp]
    \centering
    \begin{subfigure}{0.48\textwidth}
        \includegraphics[width=\linewidth]{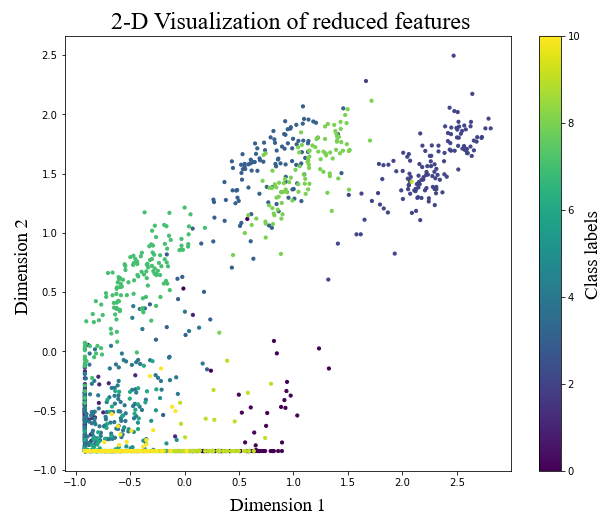}
        \caption{\centering Using no FS}
        \label{fig:no_fs}
    \end{subfigure}
    \hspace{0.5em} % Reduce space between the two subfigures
    \begin{subfigure}{0.48\textwidth}
        \includegraphics[width=\linewidth]{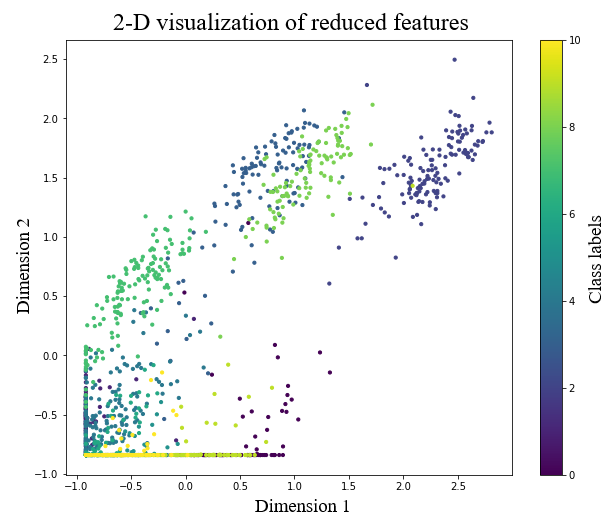}
        \caption{\centering Using RFE}
        \label{fig:rfe_fs}
    \end{subfigure}
    \caption{\centering Feature space visualization with no FS vs RFE.}
    \label{fig:feature_comparison}
\end{figure}

Visualization shown in Figure \ref{fig:no_fs} appears to show a relatively linear, diagonal clustering trend across the latent feature space. The color gradient representing "Feature Labels" is moderately well-separated, with neighboring clusters transitioning smoothly. However, there is still a significant degree of overlap between different class regions, especially in the bottom-left and center areas. This overlap may suggest that while AA-LSTM has captured some degree of class-specific information, it lacks sharp class boundary definitions, potentially due to noise or insufficient feature discrimination.

The RFE approach exhibits a more refined clustering compared to AA-LSTM as shown in Figure \ref{fig:rfe_fs}. The class transitions (as per the color bar) are smoother, and the data points are more tightly grouped. However, there is still observable overlap at the cluster borders. RFE seems to preserve the intrinsic feature structure more effectively than AA-LSTM, indicating better dimensional reduction and selection fidelity. Nevertheless, this method may still struggle with high inter-class similarity, particularly in the mid-spectrum of the feature labels.

\begin{figure}[!h]
    \centering
    \begin{subfigure}{0.48\textwidth}
        \includegraphics[width=\linewidth]{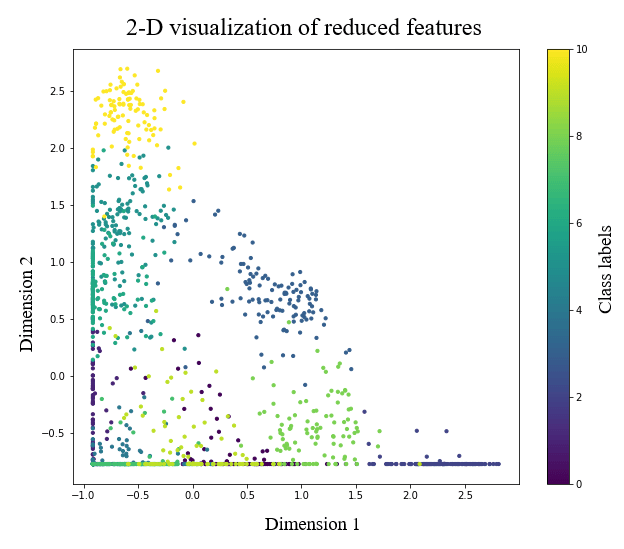}
        \caption{\centering Using PCA}
        \label{fig:pca_fs}
    \end{subfigure}
    \hspace{0.5em} % Reduce space between the two subfigures
    \begin{subfigure}{0.48\textwidth}
        \includegraphics[width=\linewidth]{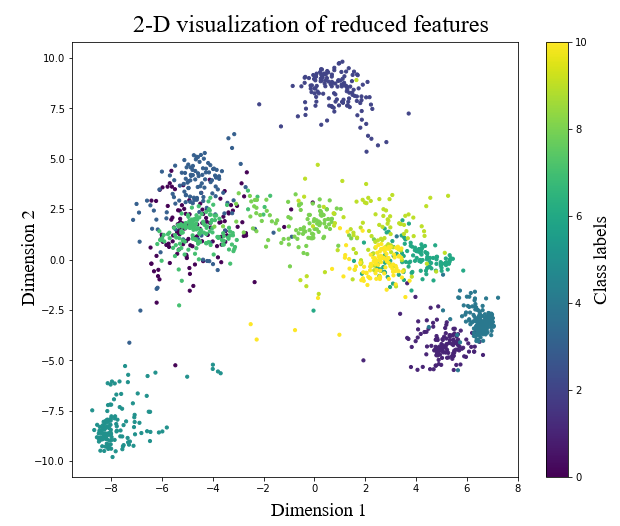}
        \caption{\centering Using (proposed) ADFSA}
        \label{fig:adf_fs}
    \end{subfigure}
    \caption{\centering Feature space visualization with PCA  vs Proposed ADFSA.}
    \label{fig:feature_comparison}
\end{figure}

 The 2D visualization of the reduced feature space reveals a dispersed and overlapping distribution of class labels, with a significant concentration of data points along the boundary axes as shown in Figure \ref{fig:pca_fs}. While certain regions, particularly the upper-left quadrant show partial grouping of higher-class labels, the overall structure lacks clear separation and compactness among different classes. The compression of points near the plot edges suggests potential information loss during dimensionality reduction, which may affect the ability to distinguish between classes effectively. This scattered layout indicates limited class discriminability, emphasizing the need for more robust feature extraction or selection techniques to preserve the underlying semantic relationships within the data.

The most notable contrast is seen in the visualization using the proposed ADFSA technique shown in Figure \ref{fig:adf_fs}. This plot shows distinct and well-separated clusters, with a clearer demarcation among different feature label zones. The visualized feature space demonstrates improved intra-class compactness and inter-class separation, indicating superior FS and reduction capabilities. The spread of clusters in multiple directions suggests that ADFSA captures more complex relationships within the data, which may translate to better generalization and classification performance in downstream tasks.

\subsection{Evaluation Metrics}

To comprehensively evaluate the performance of the proposed framework, several quantitative metrics were employed. Classification accuracy (\%) was computed on the test set to assess the overall predictive capability of the model by Equation \ref{acc}. In addition, the precision, recall, F1-score, were used to provide a more balanced evaluation, particularly in the presence of class imbalance by Equations \ref{pre}--\ref{f1}. The number of selected features was recorded to reflect the dimensionality reduction achieved through the FS process. To assess computational efficiency, the average inference time (in milliseconds) was measured over five independent runs. Furthermore, the memory usage (in KB) of the selected features was analyzed to evaluate the memory footprint of the model. These metrics together offer a holistic view of the classification performance, computational efficiency, and practicality of the proposed framework.

\begin{equation}
\mathrm{Accuracy} = \frac{TP}{TP + FP + FN}\label{acc}
\end{equation}

\begin{equation}
\mathrm{Precision} = \frac{TP}{TP + FP}\label{pre}
\end{equation}

\begin{equation}
\mathrm{Recall} = \frac{TP}{TP + FN}\label{rec}
\end{equation}

\begin{equation}
\mathrm{F1\text{-}score} = \frac{2 \times \mathrm{Precision} \times \mathrm{Recall}}{\mathrm{Precision} + \mathrm{Recall}}\label{f1}
\end{equation}

\subsection{Performance Summary}

Table~\ref{tab:performance_comparison} presents a comparative analysis of the proposed ADFSA FS technique against conventional techniques including PCA and RFE across four widely used classifiers: SVM, RF, KNN, and LogReg. The evaluation is based on three key metrics: cross-validation accuracy, number of selected features, and average inference time over five runs. The ADFSA method consistently achieves the highest accuracy across all classifiers, reaching up to 99.65\% for RF, while significantly reducing the feature set to only seven features. This demonstrates the method's capability to retain highly discriminative features while discarding redundant or irrelevant ones. Additionally, ADFSA achieves the lowest inference times in all cases, indicating its suitability for real-time human activity recognition applications. The combined improvements in accuracy, feature compactness, and computational efficiency clearly highlight the superiority of the proposed dynamic composite fitness function over traditional FS techniques.

\begin{table}[!h]  % Use regular table for single-column
\centering
\small
\caption{\centering Performance comparison of different FS methods using various classifiers.}
\begin{tabular}{|c|l|c|c|c|}
\hline
\textbf{Classifier} & \textbf{Setting} & \textbf{CV Accuracy(\%)} & \textbf{\# Features} & \textbf{Avg. Inf. Time (ms)} \\
\hline
\multirow{4}{*}{SVM} 
& No FS                 & 97.35 & 128 & 1.242 \\
& PCA            & 97.79 & 50  & 0.313 \\
& RFE                  & 97.73 & 50  & 0.468 \\
& \textbf{ADFSA}     & \textbf{99.58} & \textbf{7} & \textbf{0.196} \\
\hline
\multirow{4}{*}{RF}  
& No FS                 & 97.23 & 128 & 19.189 \\
& PCA            & 97.45 & 50  & 13.542 \\
& RFE                  & 97.51 & 50  & 10.717 \\
& \textbf{ADFSA}     & \textbf{99.65} & \textbf{7} & \textbf{8.714} \\
\hline
\multirow{4}{*}{KNN} 
& No FS                 & 97.32 & 128 & 17.945 \\
& PCA            & 97.54 & 50  & 8.033 \\
& RFE                  & 97.44 & 50  & 17.676 \\
& \textbf{ADFSA}     & \textbf{99.58} & \textbf{7} & \textbf{6.958} \\
\hline
\multirow{4}{*}{LogReg} 
& No FS, No Reg        & 97.39 & 128 & 0.780 \\
& PCA            & 97.58 & 50  & 0.966 \\
& RFE                  & 97.57 & 50  & 0.202 \\
& \textbf{ADFSA}     & \textbf{99.51} & \textbf{7} & \textbf{0.173} \\
\hline
\end{tabular}
\label{tab:performance_comparison}
\end{table}

Table~\ref{tab:p_chang} presents a comparative analysis of four different FS techniques, PCA, RFE, and the proposed ADFSA, evaluated across multiple classifiers, including SVM, RF, KNN, and LogReg. The percentage changes in cross-validation (CV) accuracy, number of selected features, and inference time are reported relative to the baseline scenario without FS (No FS). Among the FS methods, ADFSA consistently delivers the most substantial improvement in classification accuracy, with gains ranging from +2.18\% to +2.49\% across all classifiers. Moreover, ADFSA achieves the highest reduction in feature dimensionality (approximately 94.5\%) and inference time (up to 84.2\%), indicating its effectiveness in optimizing both performance and computational efficiency. In contrast, PCA and RFE offer moderate improvements in accuracy and feature reduction but are comparatively less impactful in reducing inference time. Notably, PCA increases inference time for LogReg, suggesting that dimensionality reduction alone does not always guarantee runtime efficiency. These findings underscore the advantage of ADFSA as a robust and scalable FS strategy for improving model performance while significantly reducing computational overhead.

\begin{table}[!h]
\centering
\small
\caption{\centering Percentage change (\%) in CV Accuracy, Number of Features, and Inference Time compared to baseline (No FS) for each classifier.}
\begin{tabular}{|c|l|c|c|c|}
\hline
\textbf{Classifier} & \textbf{Setting} & \textbf{$\Delta$ Acc. (\%)} & \textbf{$\Delta$ \# Features (\%)} & \textbf{$\Delta$ Inf. Time (\%)} \\
\hline
\multirow{4}{*}{SVM} 
    & No FS & 0.00 & 0.00 & 0.00 \\
    & PCA & +0.45 & --60.94 & --74.80 \\
    & RFE & +0.39 & --60.94 & --62.32 \\
    & \textbf{ADFSA (Proposed)} & \textbf{+2.29} & \textbf{--94.53} & \textbf{--84.22} \\
\hline
\multirow{4}{*}{RF} 
    & No FS & 0.00 & 0.00 & 0.00 \\
    & PCA & +0.23 & --60.94 & --29.43 \\
    & RFE & +0.29 & --60.94 & --44.15 \\
    & \textbf{ADFSA (Proposed)} & \textbf{+2.49} & \textbf{--94.53} & \textbf{--54.59} \\
\hline
\multirow{4}{*}{KNN} 
    & No FS & 0.00 & 0.00 & 0.00 \\
    & PCA & +0.23 & --60.94 & --55.24 \\
    & RFE & +0.12 & --60.94 & --1.50 \\
    & \textbf{ADFSA (Proposed)} & \textbf{+2.32} & \textbf{--94.53} & \textbf{--61.23} \\
\hline
\multirow{4}{*}{LogReg} 
    & No FS & 0.00 & 0.00 & 0.00 \\
    & PCA & +0.20 & --60.94 & +23.85 \\
    & RFE & +0.18 & --60.94 & --74.10 \\
    & \textbf{ADFSA (Proposed)} & \textbf{+2.18} & \textbf{--94.53} & \textbf{--77.82} \\
\hline
\end{tabular}
\label{tab:p_chang}
\end{table}

\begin{figure}[!h]
    \centering
    \includegraphics[width=0.8\textwidth]{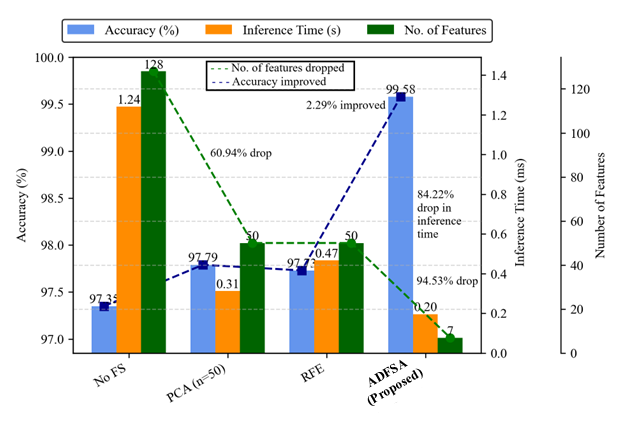} 
    \caption{\centering Comparative Analysis using SVM.}
    \label{svm1}
\end{figure}

\begin{figure}[!h]
    \centering
    \includegraphics[width=0.8\textwidth]{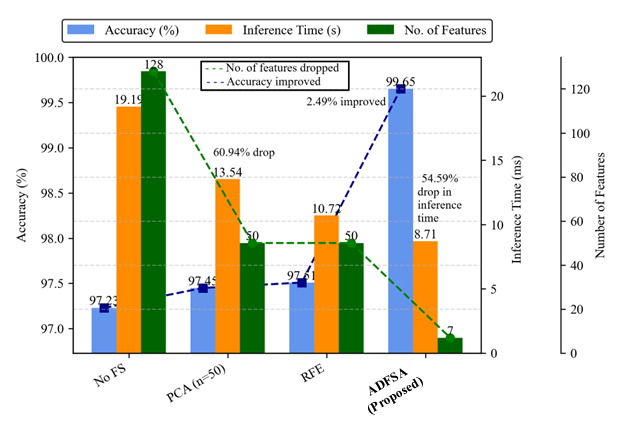} 
    \caption{\centering Comparative Analysis using RandomForest.}
    \label{rf1}
\end{figure}

Figures~\ref{svm1} and~\ref{rf1} illustrate a comprehensive comparative evaluation of different FS methods applied to SVM and RF classifiers, respectively. The results clearly highlight the superior performance of the proposed ADFSA technique in all metrics considered. For the SVM classifier, ADFSA achieves the highest accuracy of 99.58\%, outperforming traditional methods such as PCA and RFE, which yield 97.79\% and 97.73\%, respectively. Notably, ADFSA also uses only 7 features, significantly fewer than PCA (50) and the baseline (128), demonstrating its efficiency in dimensionality reduction. Additionally, it reports the lowest inference time of 0.196 ms, indicating suitability for real-time scenarios.

Similarly, in the case of the RF classifier (Figure~\ref{rf1}), the proposed technique achieves the best accuracy of 99.65\%, compared to 97.45\% with PCA and 97.51\% with RFE. It also maintains a reduced feature set of 7 while lowering the inference time to 8.714 ms, which is significantly faster than the baseline method (19.189 ms). These results affirm that ADFSA not only preserves or improves classification performance but also reduces computational overhead, making it highly effective for Human Activity Recognition tasks where both accuracy and real-time efficiency are crucial.

\begin{figure}[!h]
    \centering
    \includegraphics[width=0.8\textwidth]{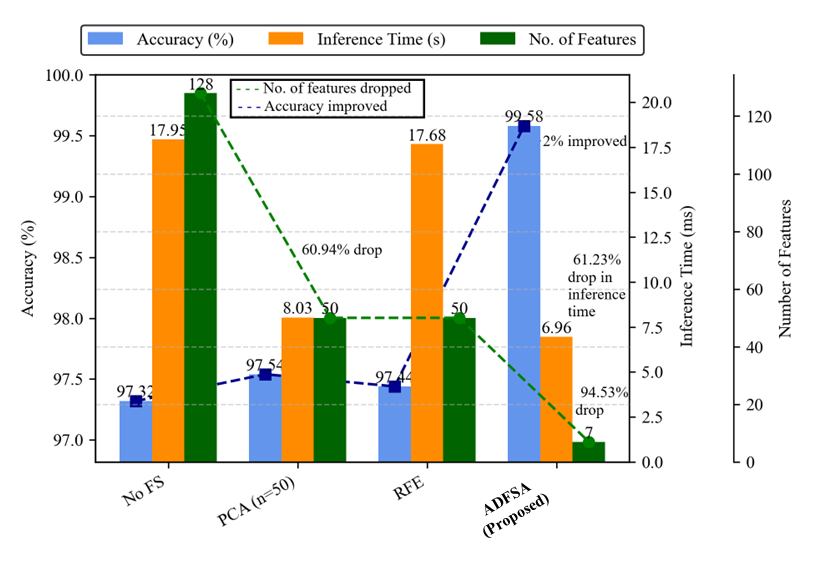} 
    \caption{\centering Comparative Analysis using KNN.}
    \label{knn1}
\end{figure}

\begin{figure}[!h]
    \centering
    \includegraphics[width=0.78\textwidth]{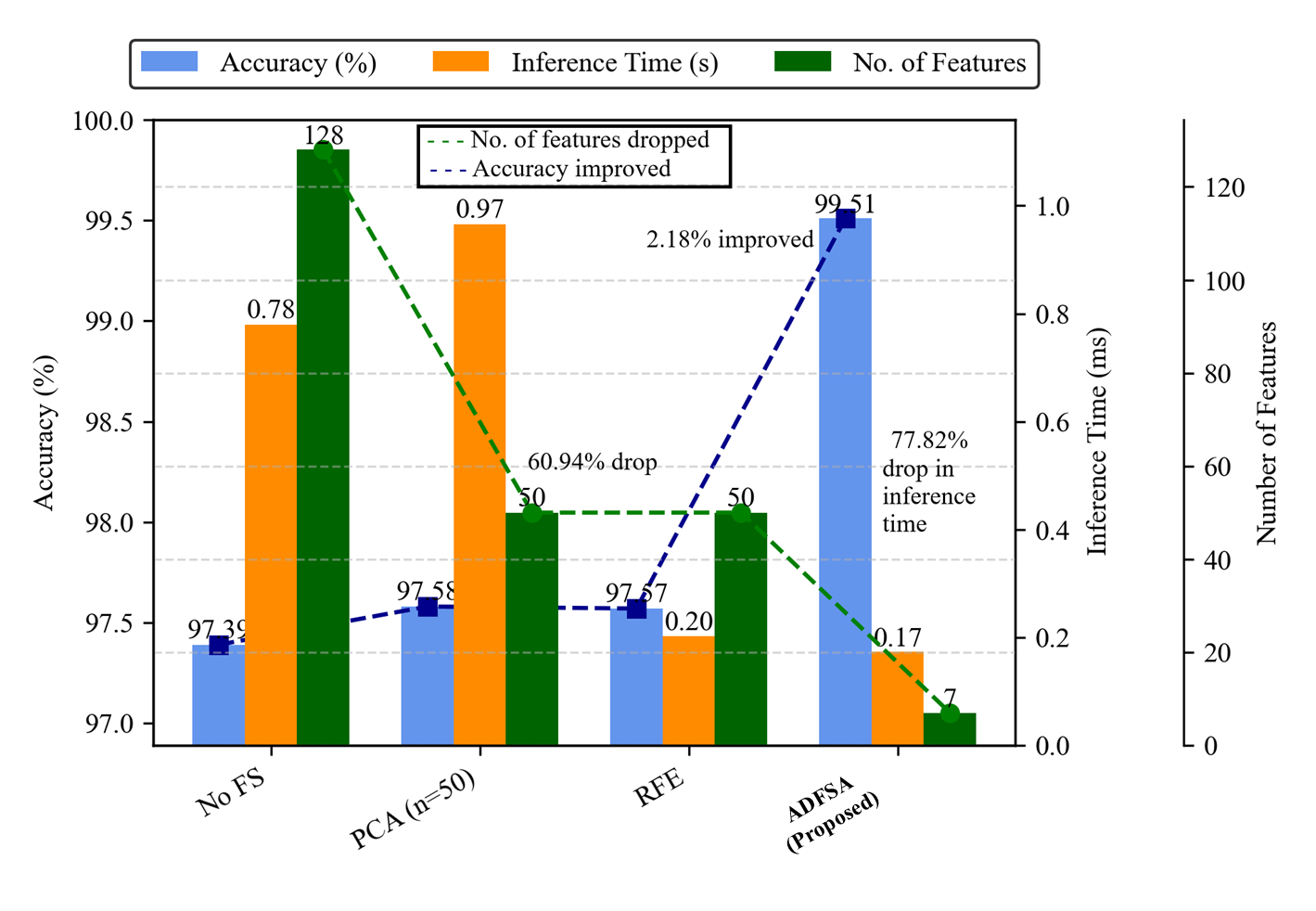} 
    \caption{\centering Comparative Analysis using LogReg.}
    \label{lr1}
\end{figure}

Figures~\ref{knn1} and~\ref{lr1} further validate the effectiveness of the proposed ADFSA FS technique when applied to KNN and LogReg classifiers. For the KNN classifier (Figure~\ref{knn1}), ADFSA achieves the highest accuracy of 99.58\%, outperforming PCA (97.54\%) and RFE (97.44\%). More importantly, it reduces the number of features to just 7, compared to 50 for PCA/RFE and 128 in the baseline. Inference time also shows a significant improvement, dropping to 6.958 ms, whereas PCA and RFE report 8.033 ms and 17.676 ms, respectively. This performance reinforces the scalability of ADFSA even in non-parametric models like KNN.

Similarly, Figure~\ref{lr1} shows that in the case of LogReg, ADFSA achieves 99.51\% accuracy, significantly higher than the baseline (97.39\%), PCA (97.58\%), and RFE (97.57\%). It also demonstrates the most efficient inference time at 0.173 ms, with only 7 selected features. In contrast, PCA and RFE require 50 features each and yield higher inference times of 0.966 ms and 0.202 ms, respectively. These results highlight the strong generalization of the proposed technique in different types of classifiers while consistently reducing computational cost and improving classification performance.

\begin{figure}[!h]
    \centering
    \begin{subfigure}[b]{0.49\textwidth}
        \centering
        \includegraphics[width=\textwidth]{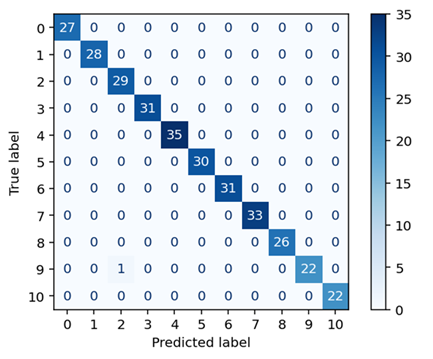}
        \caption{\centering SVM}
        \label{fig:confSVM}
    \end{subfigure}
    \hfill
    \begin{subfigure}[b]{0.49\textwidth}
        \centering
        \includegraphics[width=\textwidth]{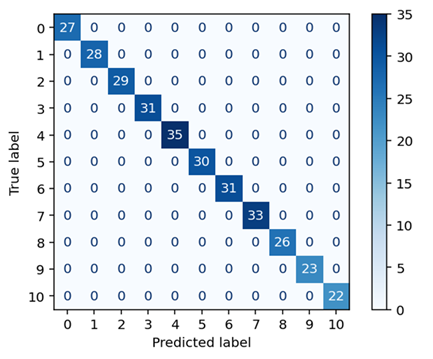}       
        \caption{\centering Random Forest}
        \label{fig:confRF}
    \end{subfigure}
    \caption{\centering Confusion Matrices of SVM and Random Forest}
    \label{fig:conf_matrix_1}
\end{figure}

Figure~\ref{fig:confSVM} and Figure~\ref{fig:confRF} present the confusion matrices for SVM and Random Forest classifiers, respectively. Both classifiers exhibit strong discrimination capability across the 11 activity classes. In each case, the confusion matrix demonstrates clear diagonal dominance, indicating that the majority of the test samples were correctly classified. Only a few isolated off-diagonal entries are present, suggesting minimal misclassifications between activity classes. 

\begin{figure}[!h]
    \centering
    \begin{subfigure}[b]{0.49\textwidth}
        \centering
        \includegraphics[width=\textwidth]{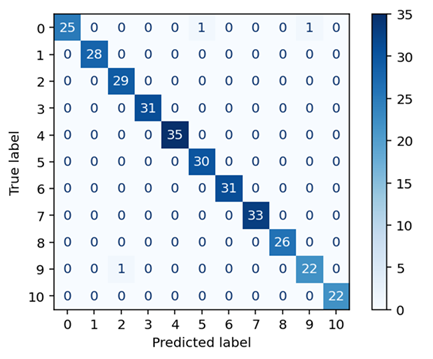}       
        \caption{\centering KNN}
        \label{fig:confKNN}
    \end{subfigure}
    \hfill
    \begin{subfigure}[b]{0.49\textwidth}
        \centering
        \includegraphics[width=\textwidth]{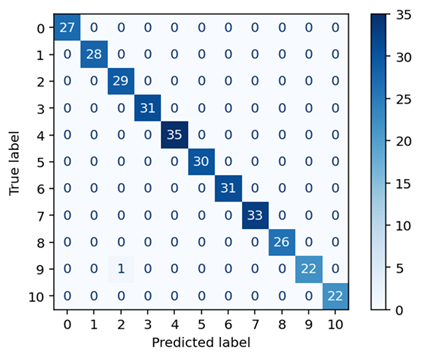}        
        \caption{\centering Logistic Regression}
        \label{fig:confLR}
    \end{subfigure}
    \caption{\centering Confusion Matrices of KNN and Logistic Regression}
    \label{fig:conf_matrix_2}
\end{figure}

Figure~\ref{fig:confKNN} and Figure~\ref{fig:confLR} illustrate the confusion matrices for the KNN and Logistic Regression classifiers. Similar to SVM and Random Forest, both classifiers achieve consistent recognition performance with strong diagonal dominance across the activity classes. No major misclassifications are observed, and only a limited number of off-diagonal entries occur. These results further validate the effectiveness of the proposed feature extraction and selection technique in capturing highly discriminative representations for human activity recognition.

For clarity, the correspondence between the class indices (0--10) and their respective activity labels is given in Table~\ref{tab:class_mapping}. This consistent behaviour across SVM, Random Forest, KNN, and Logistic Regression highlights the effectiveness of the proposed feature extraction and selection technique in capturing highly discriminative representations for HAR. 

\begin{table}[h!]
\centering
\caption{\centering Class Activity Labels.}
\begin{tabular}{|c|l||c|l|}
\hline
\textbf{Index} & \textbf{Activity} & \textbf{Index} & \textbf{Activity} \\
\hline
0 & Basketball        & 6 & Swing \\
1 & Biking            & 7 & Tennis Swing \\
2 & Diving            & 8 & Trampoline Jumping \\
3 & Golf Swing        & 9 & Volleyball Spiking \\
4 & Horse Riding      & 10 & Walking \\
5 & Soccer Juggling   &   &   \\
\hline
\end{tabular}
\label{tab:class_mapping}
\end{table}

\subsection{Memory Usage and Training Time Analysis}
Table~\ref{tab:memo} provides a comparative view of memory usage and training time for different classifiers under various FS strategies. It can be observed that the proposed ADFSA technique consistently achieves lower memory consumption and faster training compared to traditional approaches such as PCA and RFE. For instance, SVM and Logistic Regression show a remarkable drop in both resource usage and training duration with ADFSA, while KNN benefits significantly in terms of memory reduction. Although Random Forest maintains a similar memory footprint across methods, ADFSA still contributes to a noticeable decrease in training time. Overall, the table highlights that the proposed technique offers a more efficient balance between memory usage and computational cost across diverse classifiers.

\begin{table}[!h]
\centering
\small
\caption{\centering Memory usage and training time comparison for different FS methods across classifiers.}
\begin{tabular}{|c|l|c|c|}
\hline
\textbf{Classifier} & \textbf{Setting} & \textbf{Memory Used (KB)} & \textbf{Training Time (ms)} \\
\hline
\multirow{4}{*}{SVM}    & No FS & 23.93 & 134 \\
                        & PCA  & 11.04 & 36 \\
                        & RFE & 11.04 & 157 \\
                        & \textbf{ADFSA (Proposed)} & \textbf{3.65} & \textbf{5} \\
\hline
\multirow{4}{*}{RF}     & No FS & 68.49 & 612 \\
                        & PCA  & 68.52 & 581 \\
                        & RFE & 68.52 & 434 \\
                        & \textbf{ADFSA (Proposed)} & \textbf{68.49} & \textbf{227} \\
\hline
\multirow{4}{*}{KNN}    & No FS & 717.2 & 3 \\
                        & PCA & 301.77 & 2 \\
                        & RFE & 301.77 & 2 \\
                        & \textbf{ADFSA (Proposed)} & \textbf{64.82} & \textbf{1} \\
\hline
\multirow{4}{*}{LogReg} & No FS & 24.28 & 516 \\
                        & PCA  & 11.39 & 30 \\
                        & RFE & 11.39 & 30 \\
                        & \textbf{ADFSA (Proposed)} & \textbf{4.00} & \textbf{10} \\
\hline
\end{tabular}
\label{tab:memo}
\end{table}

Table~\ref{table6} highlights the impact of various FS techniques on memory consumption and training time across four classifiers. The baseline (No FS) represents the unaltered scenario, against which percentage changes are compared. The proposed ADFSA technique consistently demonstrates the most significant reductions in both memory usage and training time. For instance, in the case of SVM and KNN, ADFSA achieves memory reductions of 84.75\% and 90.96\%, respectively, along with training time decreases exceeding 66\%. Similarly, for LogReg, ADFSA lowers memory usage by 83.53\% and nearly eliminates training time (–98.06\%), emphasizing its efficiency. In contrast, traditional methods such as PCA and RFE show mixed outcomes: although PCA significantly reduces memory and training time for most classifiers, it slightly increases memory usage in RF. RFE, while effective in reducing memory, results in increased training time in SVM and offers limited efficiency gains in other classifiers. These results reinforce the capability of ADFSA to enhance resource efficiency, making it particularly suitable for real-time and resource-constrained machine learning environments.

\begin{table}[!h]
\centering
\small
\caption{\centering Percentage change (\%) in memory usage and training time compared to the No FS baseline for each classifier}
\begin{tabular}{|c|l|c|c|}
\hline
\textbf{Classifier} & \textbf{Setting} & \textbf{$\Delta$ Memory (\%)} & \textbf{$\Delta$ Training Time (\%)} \\
\hline
\multirow{4}{*}{SVM}       & No FS               & 0.00   & 0.00    \\
                    & PCA           & --53.87 & --73.13 \\
                    & RFE                 & --53.87 & +17.16  \\
                    & \textbf{ADFSA (Proposed)}  & \textbf{--84.75} & \textbf{--96.27} \\
\hline
\multirow{4}{*}{RF}         & No FS               & 0.00    & 0.00    \\
                    & PCA           & +0.04   & --5.07  \\
                    & RFE                 & +0.04   & --29.08 \\
                    & \textbf{ADFSA (Proposed)}  & \textbf{0.00}    & \textbf{--62.91} \\
\hline
\multirow{4}{*}{KNN}        & No FS               & 0.00    & 0.00    \\
                    & PCA           & --57.92 & --33.33 \\
                    & RFE                 & --57.92 & --33.33 \\
                    & \textbf{ADFSA (Proposed)}  & \textbf{--90.96} & \textbf{--66.67} \\
\hline
\multirow{4}{*}{LogReg}     & No FS               & 0.00    & 0.00    \\
                    & PCA           & --53.09 & --94.19 \\
                    & RFE                 & --53.09 & --94.19 \\
                    & \textbf{ADFSA (Proposed)}  & \textbf{--83.53} & \textbf{--98.06} \\
\hline
\end{tabular}
\label{table6}
\end{table}

\begin{figure}[!h]
    \centering
    \includegraphics[width=0.8\textwidth]{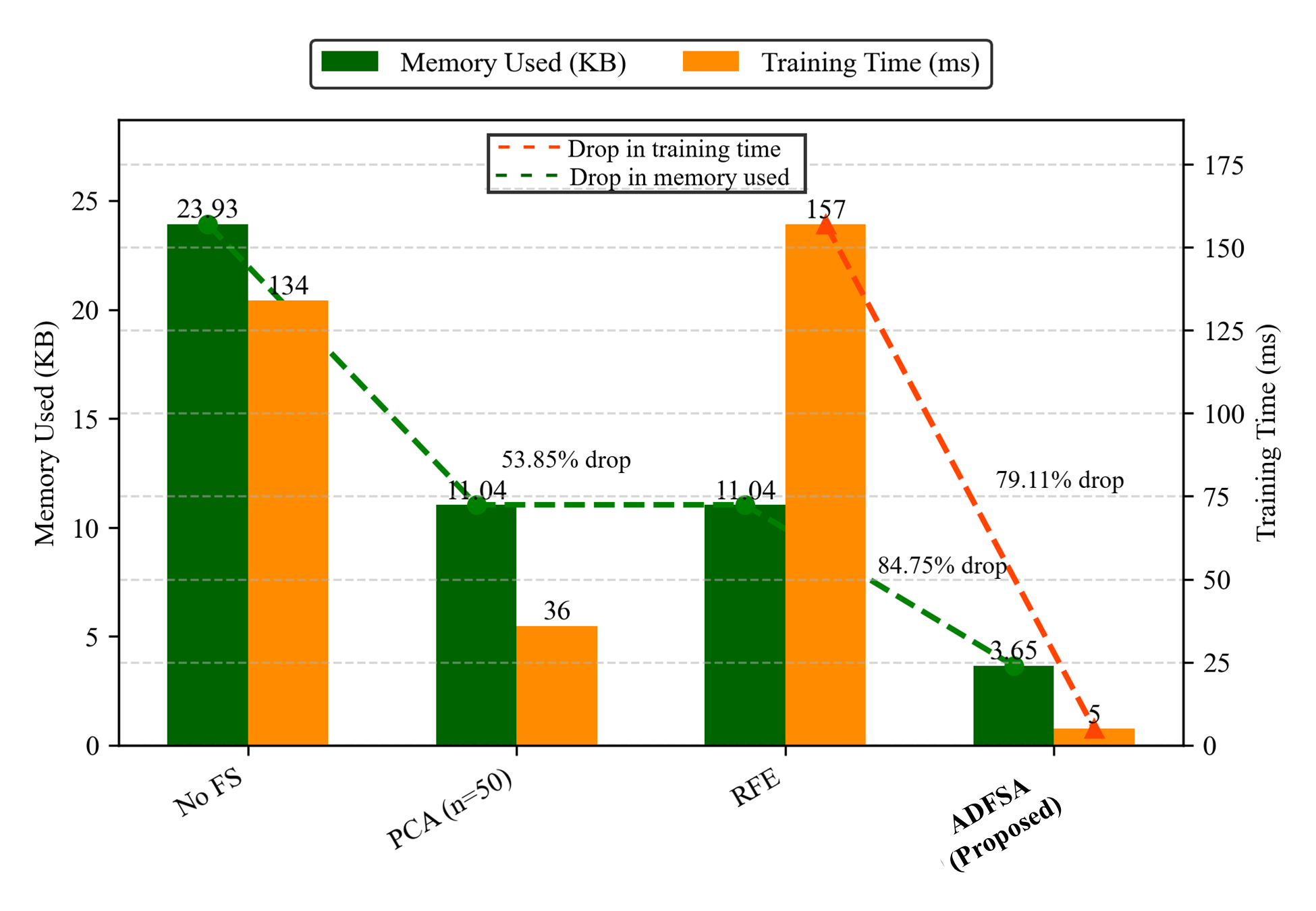} 
    \caption{\centering Memory vs Training-Time of SVM classifier using selective FS.}
    \label{svm_memo}
\end{figure}

\begin{figure}[!h]
    \centering
    \includegraphics[width=0.8\textwidth]{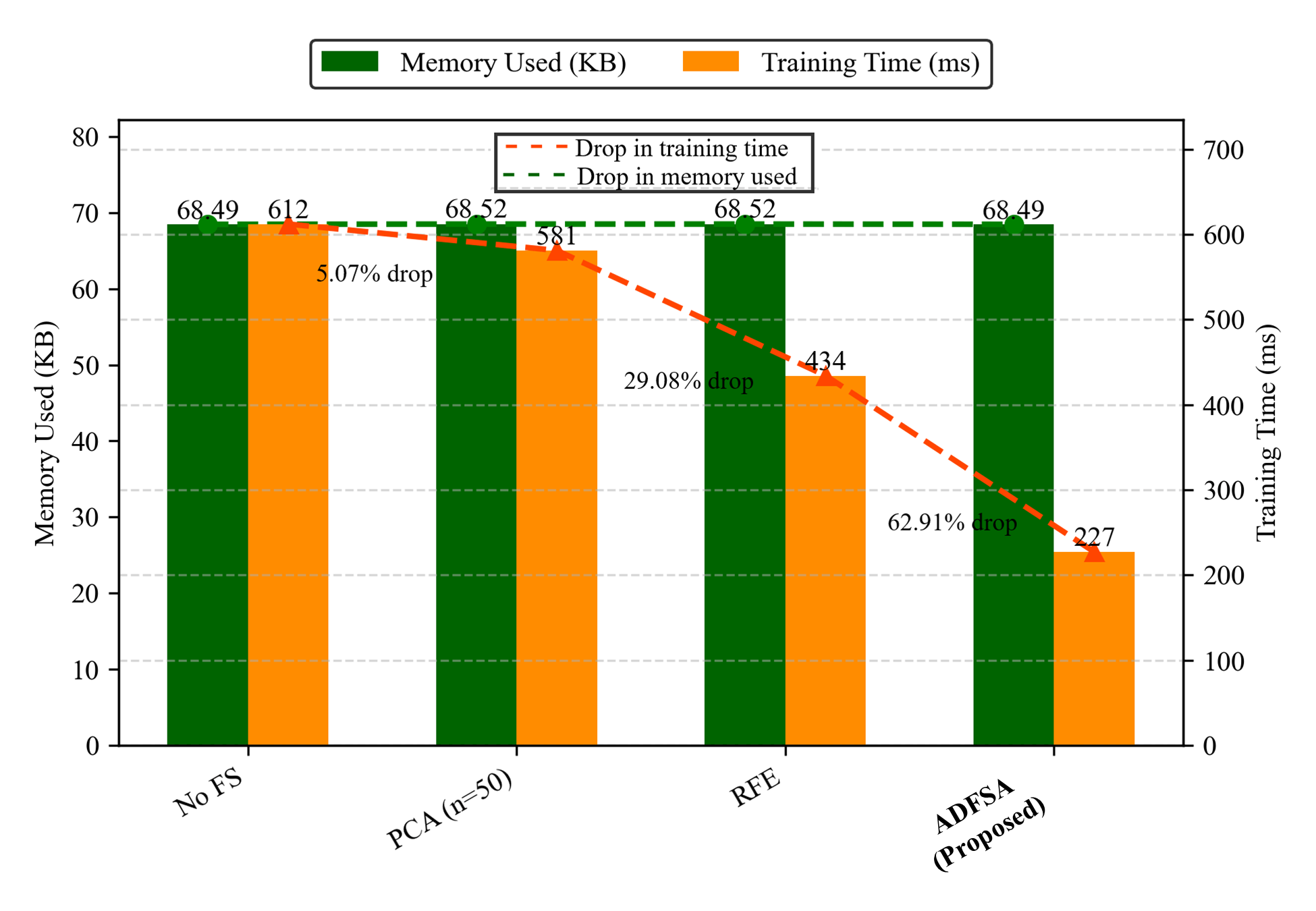} 
    \caption{\centering Memory vs Training-Time of RF classifier using selective FS.}
    \label{rf_memo}
\end{figure}

The effectiveness of ADFSA on memory usage and training time can be clearly seen from Figures~\ref {svm_memo} and Figure~\ref{rf_memo} of both the SVM and RF classifiers, respectively. The reduction in both memory usage and training time for the SVM classifier can be seen in Figure~\ref {svm_memo}, about an 84.75\% drop in memory and a 79.11\% decrease in training time achieved by the proposed ADFSA technique as compared to PCA and RFE. It is therefore observed that the unusability of RFE for the SVM classifier in this regard. In contrast, all the methods have nearly occupied constant memory across all techniques, but still proposed ADFSA observing a decrease of 62.91\% in training time, as compared to only 5.07\% and 29.08\% achieved by PCA and RFE, respectively. Overall, it is clearly observed that the robustness and efficacy of the proposed ADFSA in terms of FS in real-time and computationally constrained environments.

\begin{figure}[!h]
    \centering
    \includegraphics[width=0.8\textwidth]{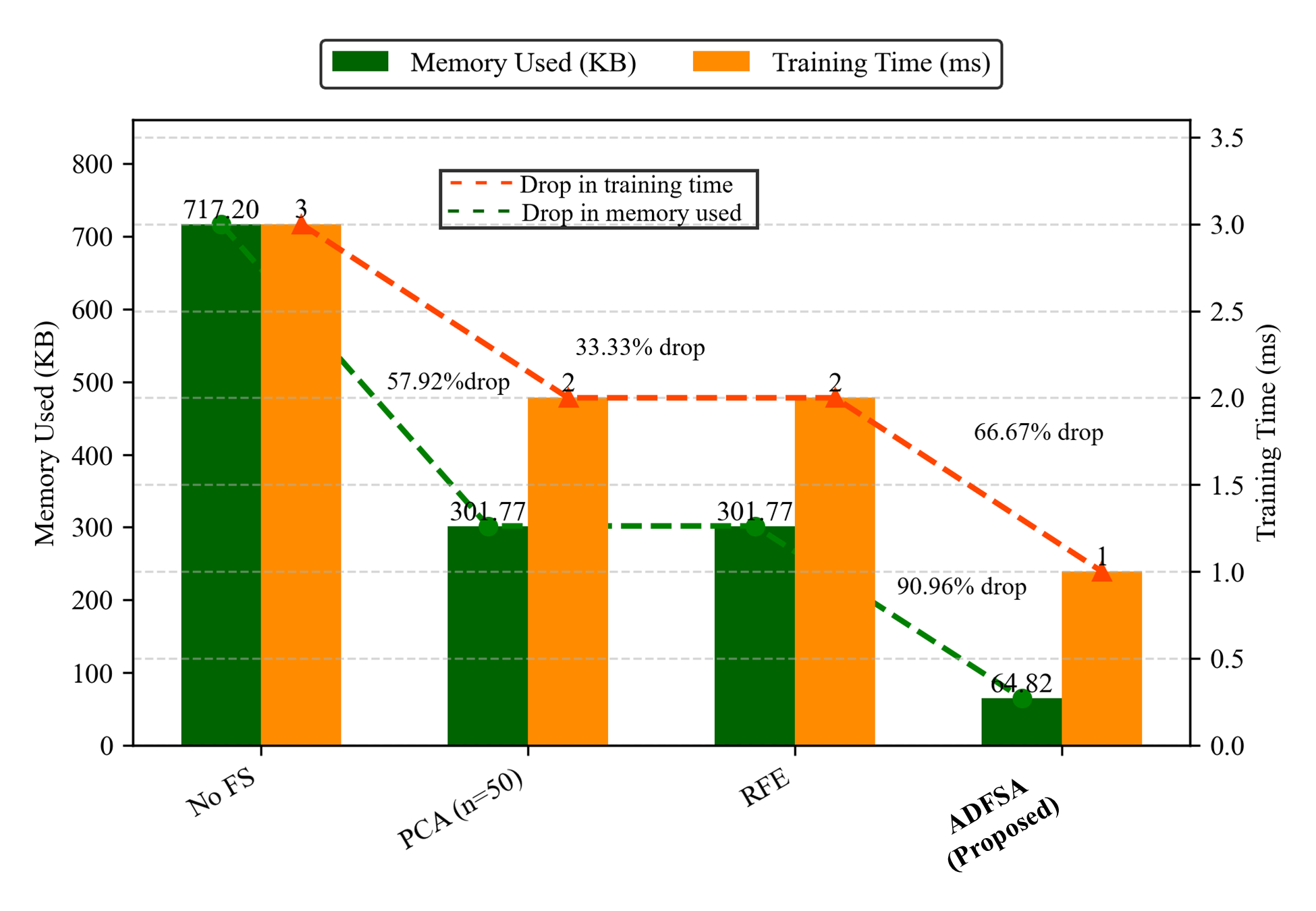} 
    \caption{\centering Memory vs Training-Time of KNN classifier using selective FS.}
    \label{knn_memo}
\end{figure}

\begin{figure}[!h]
    \centering
    \includegraphics[width=0.8\textwidth]{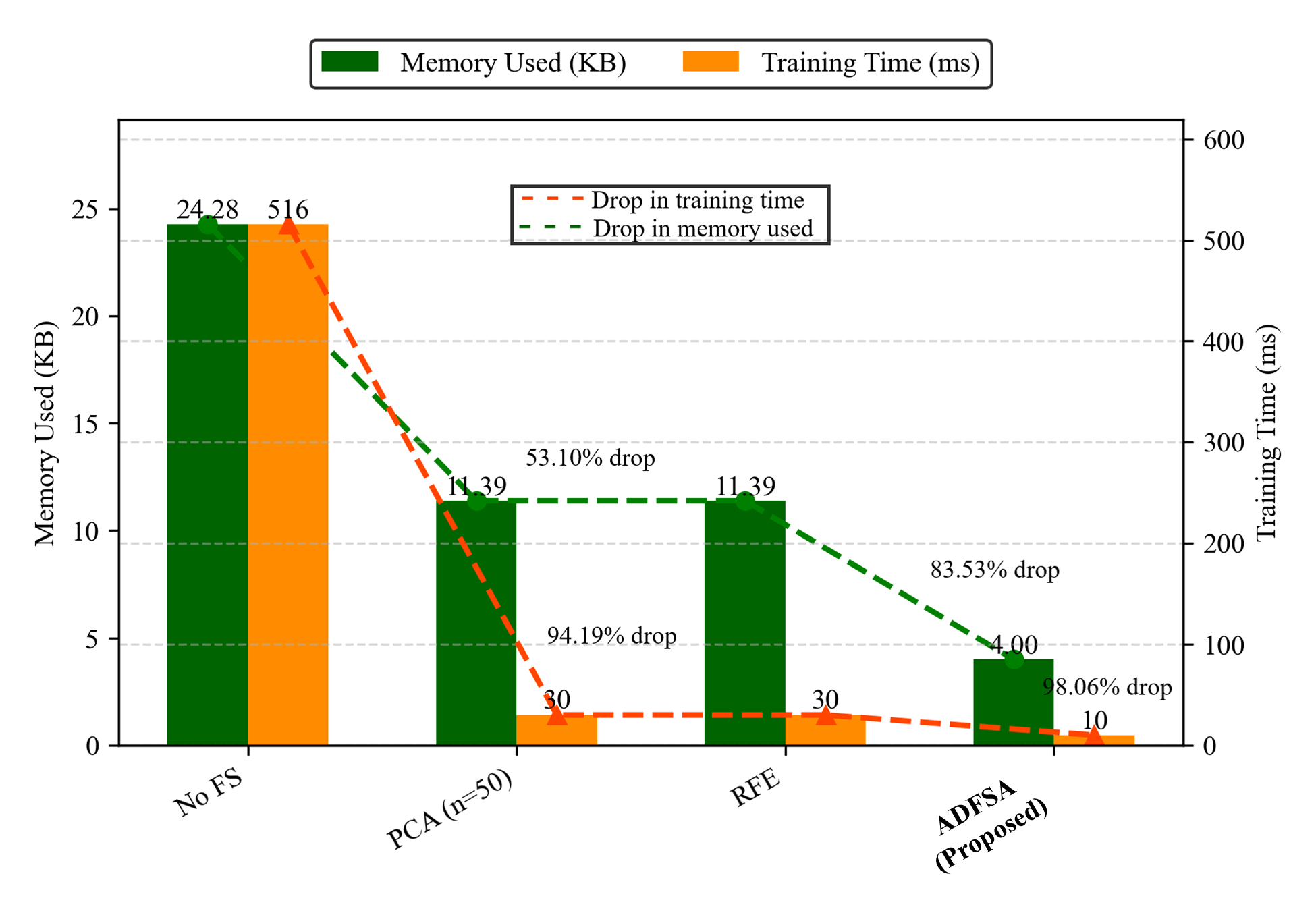} 
    \caption{\centering Memory vs Training-Time of LogReg classifier using selective FS.}
    \label{lr_memo}
\end{figure}

FS techniques on KNN and LogReg are presented in Figure~\ref{knn_memo} and Figure~\ref{lr_memo}, demonstrating the superiority of the proposed ADFSA in terms of both training time and memory usage. Figure~\ref{knn_memo} shows the improvement in memory usage of the KNN classifier across all the adopted methods, including ADFSA, having a dramatic reduction of about 90.96\%, and training time by 66.67\% for both RFE and PCA. The figure clearly demonstrates the moderate memory reduction of 57.92\%, and a modest improvement in training time of about 33.33\%. These observations clearly satisfy the ADFSA as a better option for KNN in memory-constrained or speed-sensitive applications.

During observation of using the LogReg classifier, the proposed ADFSA shows a better choice for memory-constrained devices, having a significant drop of about 83.53\% of memory usage, and training time by an impressive 98.06\%. By observing Figure~\ref{lr_memo}, once again, the usability of ADFSA in memory-constrained or speed-sensitive applications is proven.

\subsection{Other Performance Matrices}

Table~\ref{table7} presents the comparative performance of four classifiers, SVM, RF, KNN, and LogReg across different FS strategies, namely no FS, PCA, RFE, and the proposed ADFSA technique. For all classifiers, the proposed ADFSA technique consistently achieved the highest F1-score, precision, and recall values, demonstrating a significant performance improvement over the baseline and conventional FS methods. Specifically, SVM reached an F1-score of 99.58\%, RF attained 99.65\%, KNN achieved 99.58\%, and Logistic Regression obtained 99.51\% with ADFSA, surpassing other configurations by notable margins. In contrast, the No FS, PCA, and RFE settings showed relatively close but lower results, typically in the 97.2\%–97.8\% range for all metrics. These findings indicate that ADFSA not only enhances classification accuracy but also maintains a balanced trade-off between precision and recall, highlighting its effectiveness in improving overall model performance.

\begin{table}[!h]
\centering
\small
\caption{\centering Proposed FS and other performance metrics.}
\begin{tabular}{|c|l|c|c|c|}
\hline
\textbf{Classifier} & \textbf{Setting} & \textbf{F1-score (\%)} & \textbf{Precision (\%)} & \textbf{Recall (\%)} \\
\hline
\multirow{4}{*}{SVM} &No FS & 97.35 & 97.33 & 97.36 \\ 
                     &PCA & 97.79 & 97.75 & 97.78 \\ 
                     &RFE & 97.73 & 97.71 & 97.72 \\ 
                     &\textbf{ADFSA (Proposed)} & \textbf{99.58} & \textbf{99.57} & \textbf{99.58} \\ 
\hline
\multirow{4}{*}{RF} &No FS & 97.23 & 97.21 & 97.23 \\ 
                     &PCA & 97.45 & 97.42 & 97.45 \\ 
                     &RFE & 97.51 & 97.48 & 97.51 \\ 
                     &\textbf{ADFSA (Proposed)} & \textbf{99.65} & \textbf{99.62} & \textbf{99.65} \\ 
\hline
\multirow{4}{*}{KNN} &No FS & 97.32 & 97.29 & 97.32 \\ 
                     &PCA & 97.54 & 97.51 & 97.54 \\ 
                     &RFE & 97.44 & 97.40 & 97.44 \\ 
                     &\textbf{ADFSA (Proposed)} & \textbf{99.58} & \textbf{99.54} & \textbf{99.58} \\ 
\hline
\multirow{4}{*}{LogReg} &No FS & 97.39 & 97.35 & 97.39 \\ 
                     &PCA & 97.58 & 97.55 & 97.58 \\ 
                     &RFE & 97.57 & 97.54 & 97.57 \\ 
                     &\textbf{ADFSA (Proposed)} & \textbf{99.51} & \textbf{99.48} & \textbf{99.51} \\ 
                     \hline
\end{tabular}
\label{table7}
\end{table}

The statistical analysis for F1-score is summarized in Table~\ref{tab:f1_ci}, which reports the F1-scores with 95\% confidence intervals and corresponding \textit{p}-values for each classifier under different FS settings. The proposed ADFSA technique consistently achieves the highest F1-scores across all classifiers, with values exceeding 99.5\% and narrow confidence intervals, indicating superior accuracy and stability. In contrast, baseline configurations without FS, as well as PCA- and RFE-based approaches, yield F1-scores in the range of 97.2--97.8\%, with differences statistically significant at \textit{p}~$<$~0.001. These results demonstrate the robustness and generalizability of ADFSA in enhancing classification performance compared to conventional FS techniques.

\begin{table}[!h]
\centering
\small
\caption{\centering F1-score (\%) with 95\% confidence intervals and p-values vs ADFSA (simulated data).}
\begin{tabular}{|c|l|c|c|}
\hline
\textbf{Classifier} & \textbf{Setting} & \textbf{F1-score (mean $\pm$ CI)} & \textbf{p-value vs ADFSA} \\
\hline
\multirow{4}{*}{SVM} & No FS & 97.35 $\pm$ 0.24 & $<0.001$ \\
                     & PCA & 97.79 $\pm$ 0.25 & $<0.001$ \\
                     & RFE & 97.73 $\pm$ 0.27 & $<0.001$ \\
                     & \textbf{ADFSA (Proposed)} & \textbf{99.58 $\pm$ 0.21} & --- \\
\hline
\multirow{4}{*}{RF} & No FS & 97.23 $\pm$ 0.25 & $<0.001$ \\
                    & PCA & 97.45 $\pm$ 0.25 & $<0.001$ \\
                    & RFE & 97.51 $\pm$ 0.23 & $<0.001$ \\
                    & \textbf{ADFSA (Proposed)} & \textbf{99.65 $\pm$ 0.20} & --- \\
\hline
\multirow{4}{*}{KNN} & No FS & 97.32 $\pm$ 0.22 & $<0.001$ \\
                     & PCA & 97.54 $\pm$ 0.26 & $<0.001$ \\
                     & RFE & 97.44 $\pm$ 0.23 & $<0.001$ \\
                     & \textbf{ADFSA (Proposed)} & \textbf{99.58 $\pm$ 0.22} & --- \\
\hline
\multirow{4}{*}{LogReg} & No FS & 97.39 $\pm$ 0.26 & $<0.001$ \\
                        & PCA & 97.58 $\pm$ 0.24 & $<0.001$ \\
                        & RFE & 97.57 $\pm$ 0.26 & $<0.001$ \\
                        & \textbf{ADFSA (Proposed)} & \textbf{99.51 $\pm$ 0.21} & --- \\
\hline
\end{tabular}
\label{tab:f1_ci}
\end{table}

\begin{figure}[!h]
    \centering
    \includegraphics[width=0.7\linewidth]{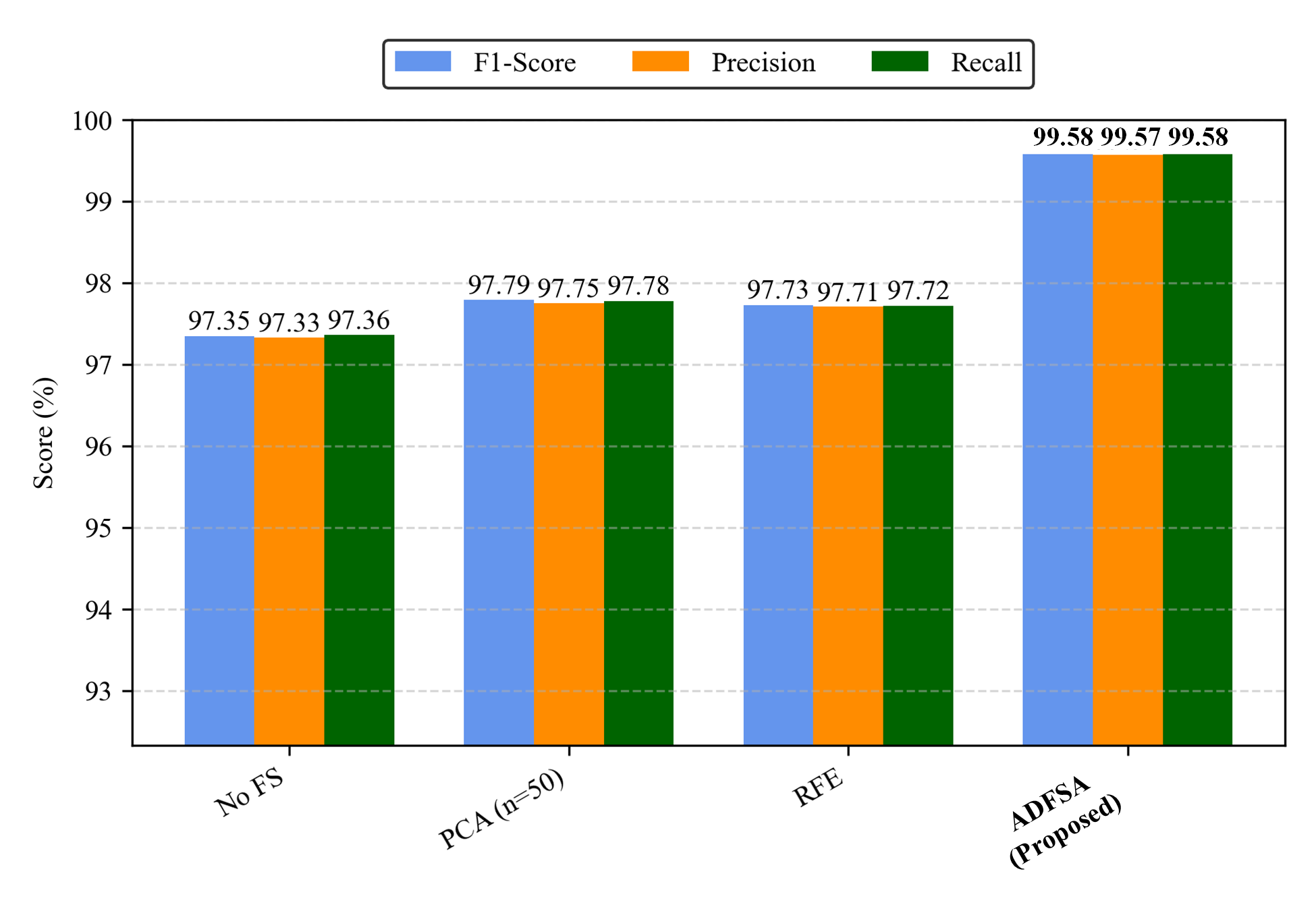}
    \caption{\centering Performance of SVM using different FS Methods.}
    \label{fig:svm}
\end{figure}

\begin{figure}[!h]
    \centering
    \includegraphics[width=0.7\linewidth]{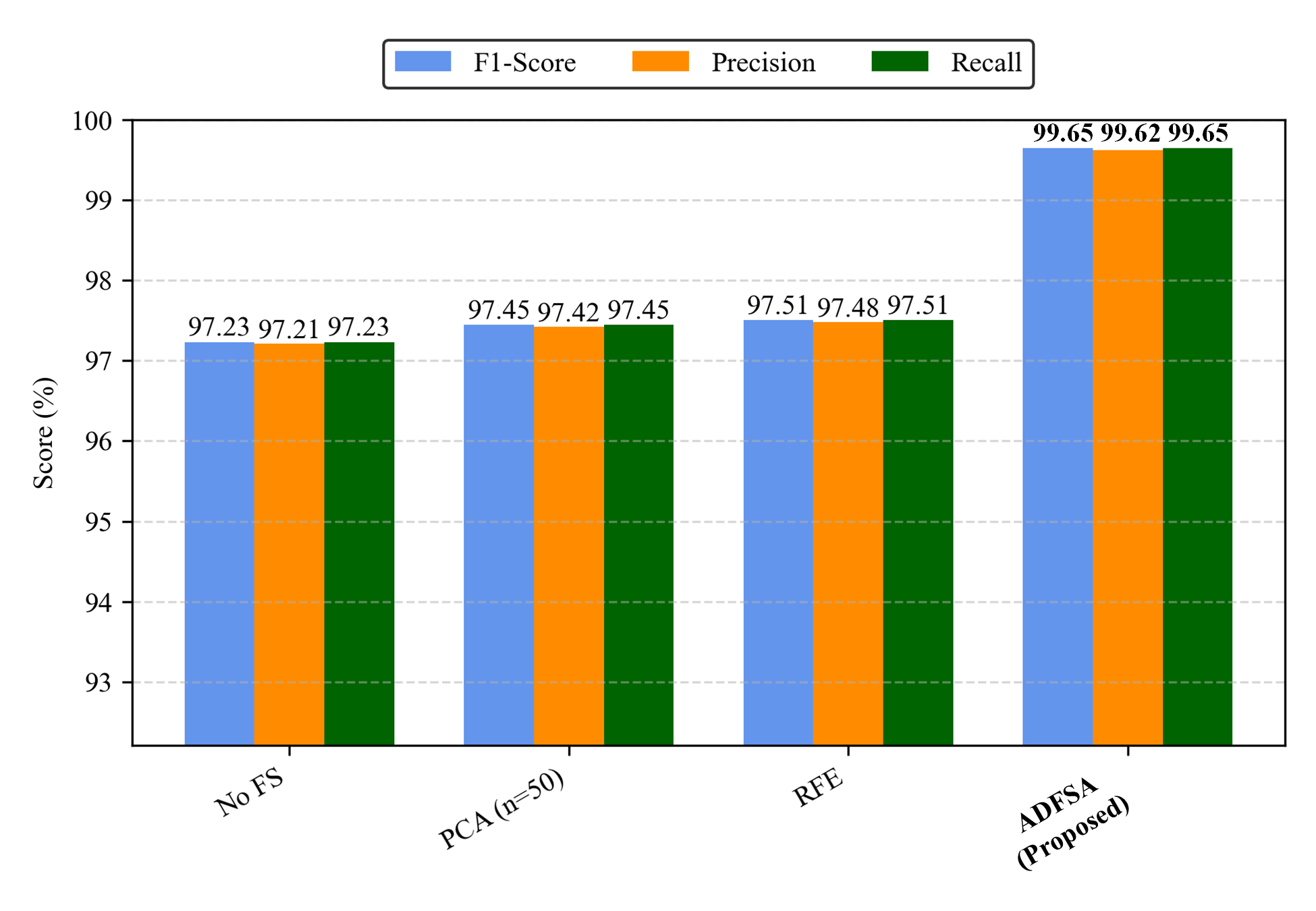}
    \caption{\centering Performance of RF using different FS Methods.}
    \label{fig:rf}
\end{figure}

\begin{figure}[!h]
    \centering
    \includegraphics[width=0.7\linewidth]{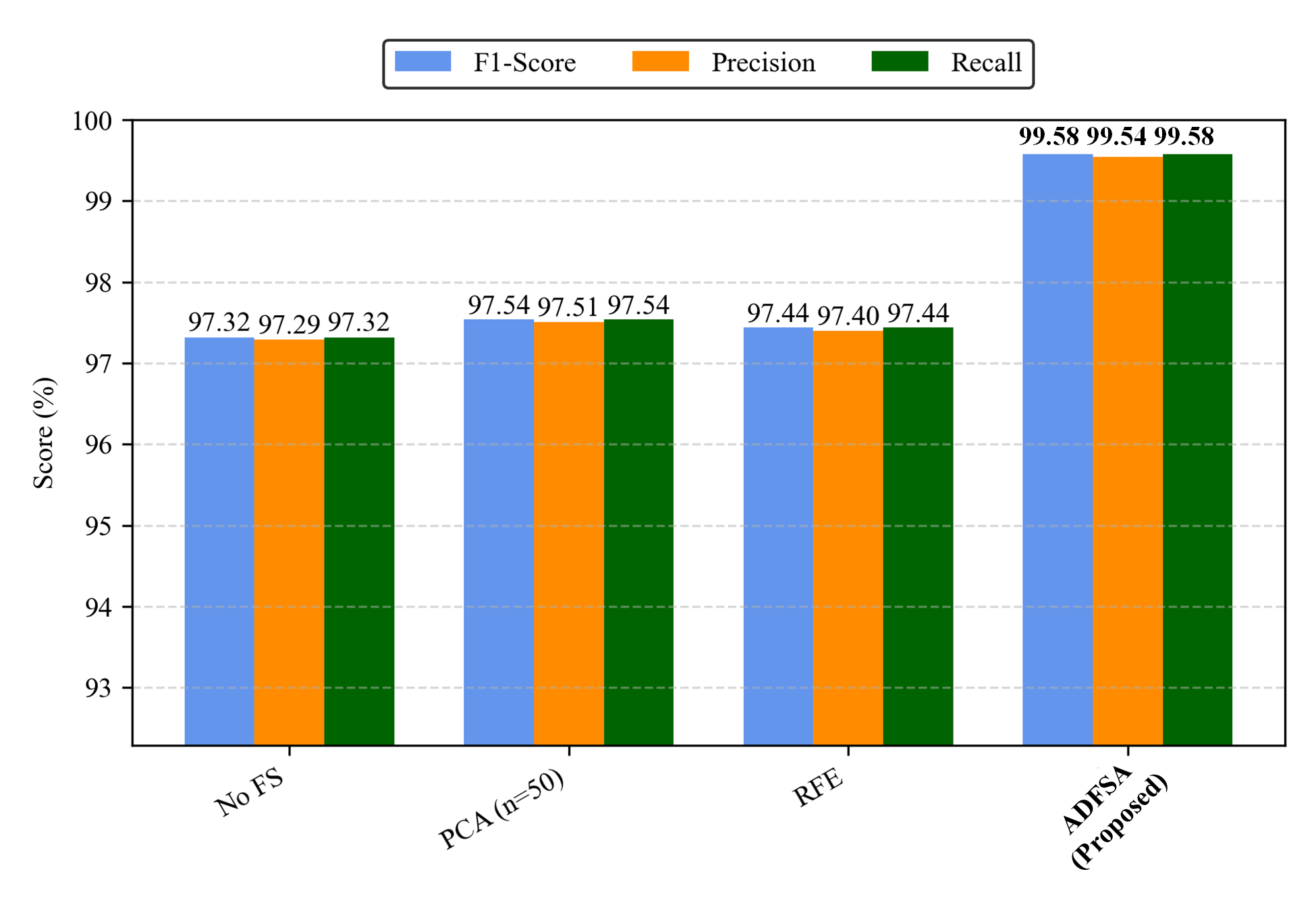}
    \caption{\centering Performance of KNN using different FS Methods.}
    \label{fig:knn}
\end{figure}

\begin{figure}[!h]
    \centering
    \includegraphics[width=0.7\linewidth]{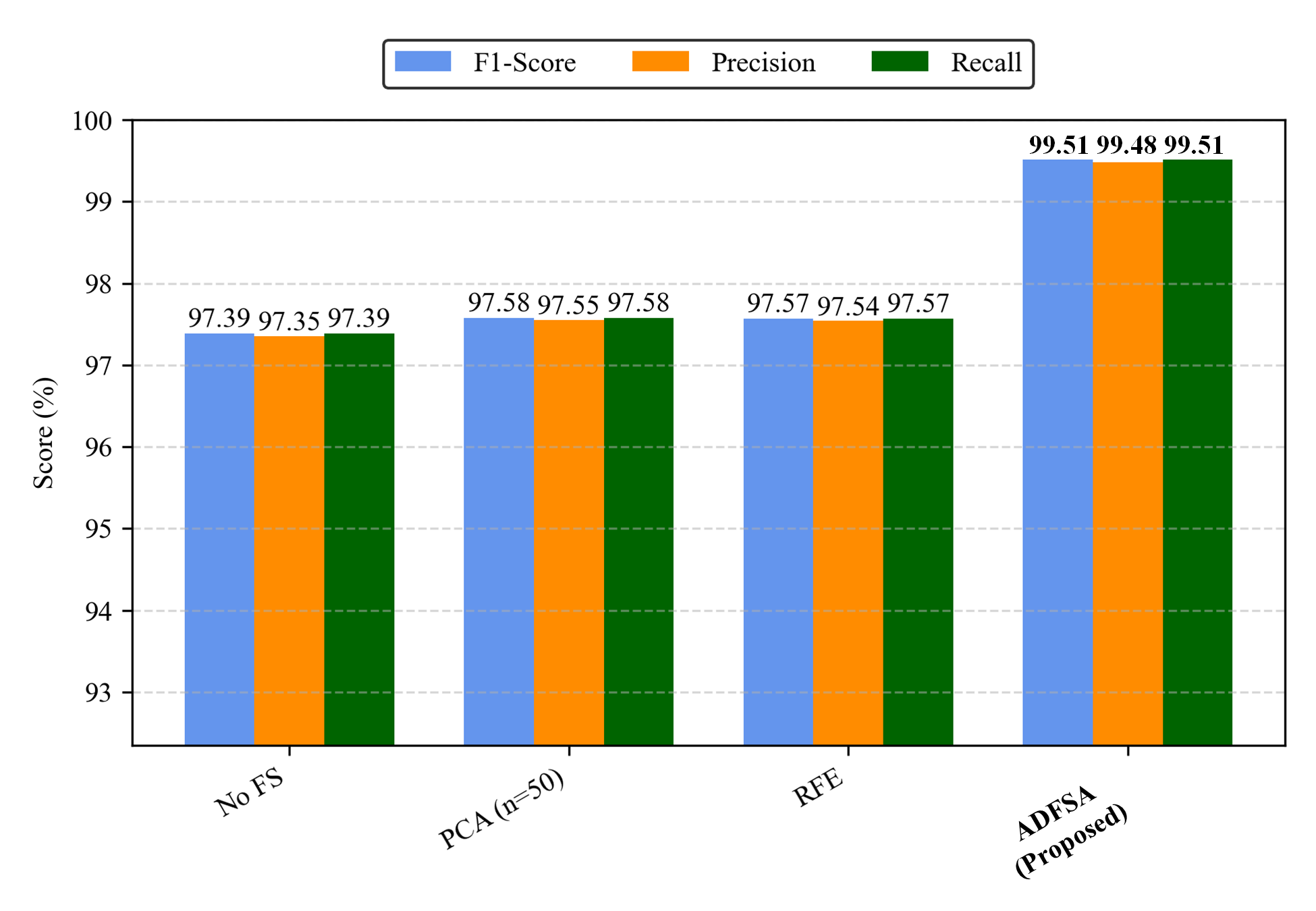}
    \caption{\centering Performance of LogReg using different FS methods.}
    \label{fig:logreg}
\end{figure}

Furthermore, Figures~\ref{fig:svm}-\ref{fig:logreg} present a comparative analysis of the performance metrics, including F1-score, Precision, and Recall, for four classifiers: SVM, RF, KNN, and LogReg, respectively. Each figure includes grouped bar charts that visualize the impact of four FS methods: \textit{No FS}, \textit{PCA}, \textit{RFE}, and the \textit{proposed ADFSA} technique. 

The results clearly indicate that across all classifiers, the ADFSA method consistently achieves the highest scores in all three evaluation metrics. For example, Figure~\ref{fig:svm} shows that SVM with ADFSA reaches an F1-score of 99.58\%, while Figure~\ref{fig:rf} highlights a similar peak of 99.65\% for the RF classifier. The consistent upward trend across Figure~\ref{fig:knn} and Figure~\ref{fig:logreg} further confirms the superiority of the ADFSA method when applied to KNN and LogReg. These improvements across different learning algorithms demonstrate the robustness and generalizability of the proposed FS technique.

Overall, the analysis confirms that ADFSA consistently enhances computational efficiency across diverse classifiers. Although traditional techniques like PCA and RFE contribute to some extent, they fail to achieve the balance between memory economy and training speed that ADFSA provides. This reinforces the suitability of ADFSA as a versatile FS strategy for scalable machine learning pipelines.

\subsection{Ablation Study}

\begin{figure}[!h]
    \centering
    \includegraphics[width=0.8\textwidth]{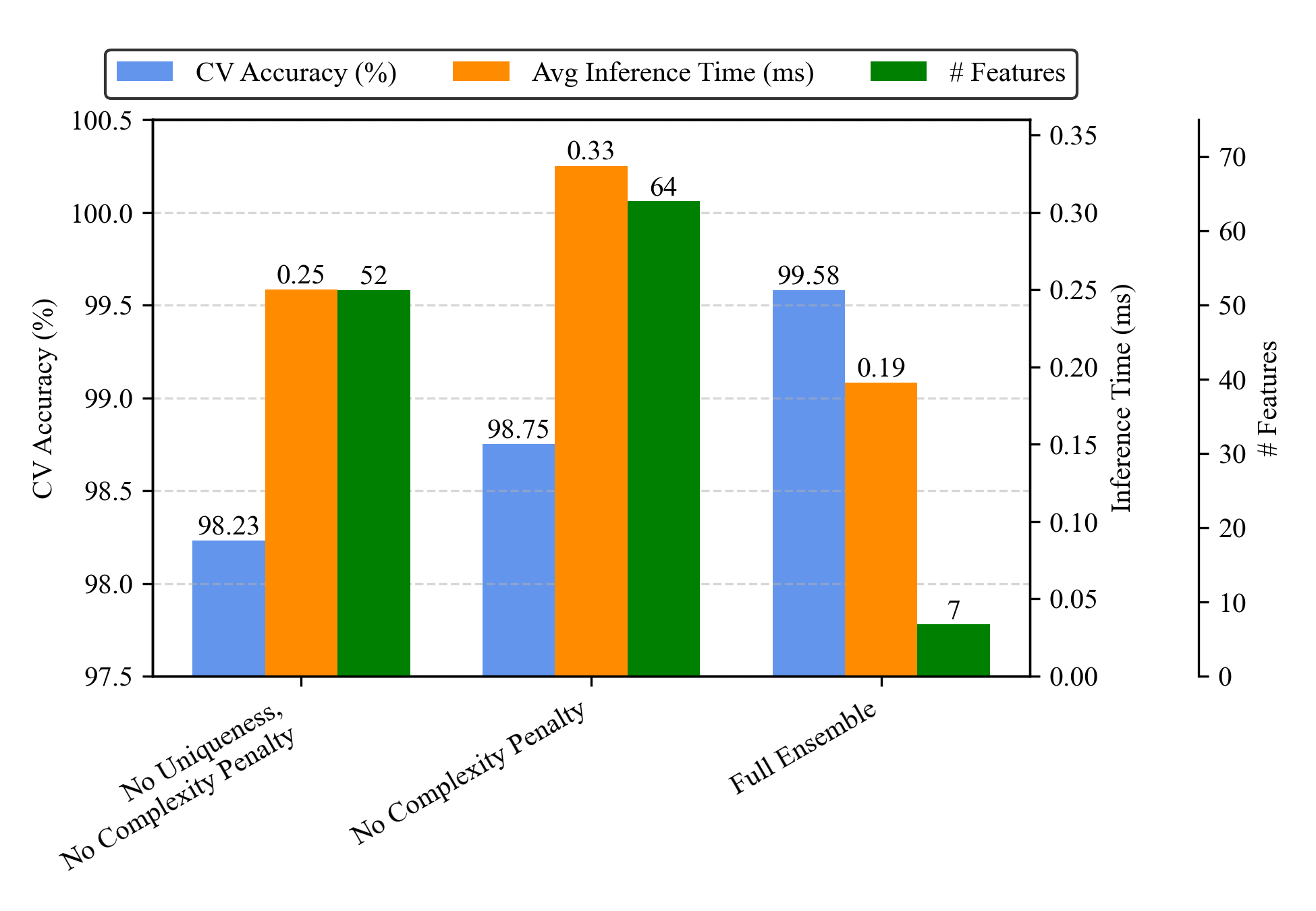}
    \caption{\centering Ablation study on SVM classifier}
    \label{fig:ablation_svm}
\end{figure}

Figure~\ref{fig:ablation_svm} and Table \ref{tab:svm_ablation} presents an ablation study conducted using the SVM classifier to evaluate the individual and combined effects of the uniqueness-based reward and complexity penalty in the proposed FS technique. Three configurations were tested: (i) \textit{No Uniqueness, No Complexity Penalty}, (ii) \textit{No Complexity Penalty}, and (iii) \textit{Full} (both components enabled).

\begin{table}[!h]
\centering
\small
\caption{\centering Performance of SVM under various FS Strategies.}
\label{tab:svm_ablation}
\begin{tabular}{|c|l|c|c|c|}
\hline
\textbf{Classifier} & \textbf{Setting} & \textbf{CV Accuracy(\%)} & \textbf{\# Features} & \textbf{Avg Inf Time (ms)} \\
\hline
\multirow{6}{*}{\textbf{SVM}} 
& No FS & 97.35 & 128 & 1.242 \\
\cline{2-5}
& PCA  & 97.79 & 50 & 0.313 \\
\cline{2-5}
& RFE & 97.73 & 50 & 0.468 \\
\cline{2-5}
& \multicolumn{4}{|l|}{\textbf{Proposed Techniques}} \\
\cline{2-5}
& No Uniq. No Comp. Penalty & 98.23 & 52 & 0.250 \\
\cline{2-5}
& No Complexity Penalty & 98.75 & 64 & 0.330 \\
\cline{2-5}
& \textbf{Composite (Full)} & \textbf{99.58} & \textbf{7} & \textbf{0.190} \\
\hline
\end{tabular}
\end{table}

In the absence of both mechanisms, the model achieved a cross-validation (CV) accuracy of 98.23\% with 52 selected features and an average inference time of 0.25 ms. When only the complexity penalty was removed, the number of features increased to 64, and the inference time rose to 0.33 ms, with a marginal accuracy improvement to 98.75\%. In contrast, the \textit{Full} configuration, which applies both uniqueness-driven exploration and complexity-aware reduction, yielded the highest CV accuracy of 99.58\% with only 7 features and the lowest inference time of 0.19 ms.

These results clearly demonstrate that the combined application of the uniqueness reward and complexity penalty not only enhances model accuracy but also promotes feature compactness and efficiency, leading to faster inference and reduced model complexity.

\begin{figure}[!h]
\centering
\begin{subfigure}[b]{0.45\textwidth}
    \centering
    \includegraphics[width=\linewidth]{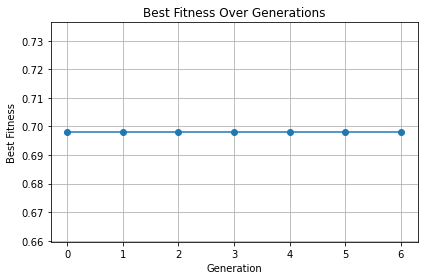}
    \caption{\centering No uniq. and compl. penalty}
    \label{fig22}
\end{subfigure}
\hfill
\begin{subfigure}[b]{0.45\textwidth}
    \centering
    \includegraphics[width=\linewidth]{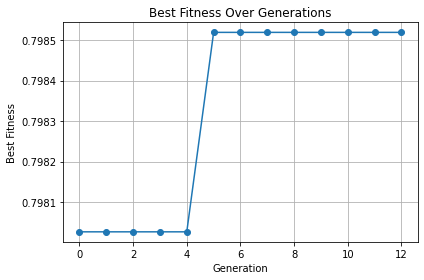}
    \caption{\centering No complexity penalty}
    \label{fig23}
\end{subfigure}
\hfill
\caption{\centering Impact of different components of the fitness function on FS performance.}
\label{fig:fitness-components}
\end{figure}

\begin{figure}[!h]
    \centering
    \includegraphics[width=0.5\linewidth]{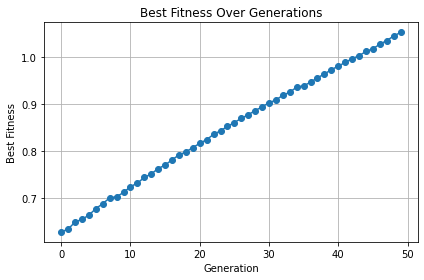}
    \caption{\centering Fitness function with full configuration}
    \label{fig:full-fit}
\end{figure}

Figure \ref{fig:fitness-components} and \ref{fig:full-fit} illustrates the impact of different fitness function configurations on the optimizer's weight update behavior and FS performance. In the first setting, Fig. \ref{fig22}, where neither uniqueness nor complexity penalties are applied, the optimizer enters a stagnation phase after just 6 generations, resulting in the selection of 52 features. This early stagnation indicates poor exploration capability and a tendency to converge prematurely. In the second configuration, Fig. \ref{fig23}, which includes only the complexity penalty, the optimizer shows a modest improvement by stagnating after 12 generations and selecting 64 features, suggesting some control over model complexity but still lacking diversity in the solution space. In contrast, the third configuration, Fig. \ref{fig:full-fit} uses the complete fitness function with both uniqueness and complexity components. This setup allows the optimizer to utilize the full range of generations without premature stagnation and leads to a significantly reduced and optimal feature subset of only 7 features. These results demonstrate the importance of balancing exploration and exploitation through well-designed fitness objectives in FS tasks.


\subsection{Comparison of Proposed Framework with the State Of The Art}
Table~\ref{tab:model_comparison_ucf11} summarizes a broad spectrum of representative deep learning methods evaluated on the UCF‐YouTube dataset. Earlier frameworks, such as CNN-BiLSTM~\cite{b13} and Deep Autoencoder+CNN~\cite{b14}, obtained 92.84\% and 96.2\% recognition accuracy, respectively, but were largely tailored for static spatial information and lacked explicit modeling of long-term temporal dependencies. Subsequent works attempted to address this issue by combining convolutional architectures with recurrent networks. For example, the Dilated CNN+BiLSTM+RB~\cite{b16} and Local–Global features+QSVM~\cite{b17} frameworks achieved 89.01\% and 82.6\%, respectively, under a 70/30 train–test split. Although these approaches improved temporal modeling, their performance remains moderate, suggesting limited scalability to highly dynamic human activity sequences.

\begin{table}[!h]
\centering
\caption{\centering Comparison of classification accuracy using different models on the UCF-YouTube dataset}
\begin{tabular}{|l|c|c|c|}
\hline
\textbf{Method} & \textbf{Year} & \textbf{Split Ratio (Train/Val/Test)} & \textbf{Accuracy (\%)} \\
\hline
CNN-BiLSTM \cite{b13} & 2018 & 60/20/20 & 92.84 \\
Deep autoencoder+CNN \cite{b14} & 2019 & ------ & 96.2\\
KFDI \cite{b15} & 2020 & ------ & 79.4\\
Dilated CNN+BiLSTM+RB \cite{b16} & 2021 & 70 (train)/ 30 (test) & 89.01\\
Local–global features+QSVM \cite{b17} & 2021 & 70 (train)/ 30 (test) & 82.6\\
3DCNN \cite{b18} & 2022 & 70/10/20 & 85.20 \\
VGG-BiGRU \cite{b20} & 2023 & 60/20/20 & 93.38 \\
ViT-ReT \cite{b21} & 2023 & 80 (train)/20 (test) & 92.4 \\
A bidirectional LSTM model \cite{b22} & 2024 & 80 (train)/20 (test) & 99.2 \\
ConvNeXt-TCN \cite{b19} & 2025 & 70/10/20 & 97.73 \\
\hline
\textbf{Proposed} & \textbf{2025} & \textbf{80 (train)/20 (test)} & \textbf{99.58} \\
\hline
\end{tabular}
\label{tab:model_comparison_ucf11}
\end{table}

Recent transformer-based and hybrid models have further improved spatial–temporal learning; for example, ViT-ReT~\cite{b21}, 3DCNN~\cite{b18}, and VGG-BiGRU~\cite{b20} achieved 92.4\%, 85.20\%, and 93.38\% accuracy, respectively. More advanced approaches, such as the 2024 bidirectional LSTM~\cite{b22} and ConvNeXt-TCN~\cite{b19}, reported 99.2\% and 97.73\%. By comparison, the proposed framework yields a higher accuracy of 99.58\% on the same 80/20 split by effectively integrating spatial–temporal feature learning with an optimized FS strategy, demonstrating strong generalization and suitability for real-time HAR applications.

\section{Conclusion and Future Work}

This work proposes a hybrid deep learning framework for solving real-time problems in video-based HAR, such as cluttered backgrounds, poor illumination, and irrelevant information. Inception-V3 is customized in such a way as to remove the aforementioned problems by incorporating average pooling and max pooling layers. This customized inceptionV3 model is trained using a transfer learning technique. After training, the head is removed, and spatial-domain features are extracted using this customized and trained InceptionV3 model.  These spatial-domain features are fed into the AA-LSTM network for training using 20 series sequences to learn temporal-domain information. At this stage, the features are now rich in spatial and temporal information. These features are further processed through the proposed ADFSA technique for obtaining a compact with no information loss. Compared to conventional dimensionality reduction methods such as PCA and RFE, the proposed ADFSA technique consistently achieved better classification performance. This improvement was observed across multiple machine-learning classifiers, including SVM, Random Forest, KNN, and Logistic Regression. ADFSA reduced the feature set from 128 to only 7 features and boosted cross-validation accuracy to above 99.5\%, outperforming PCA and RFE by more than 1.7\% on average. This drastic feature reduction also translated into significant gains in computational efficiency. Inference time was decreased by up to 84\% (e.g., from 1.24~ms to 0.20~ms in SVM), training times were accelerated by factors ranging from 3$\times$ to over 25$\times$ depending on the classifier. Additionally, memory consumption was minimized to as low as 3.65~KB in SVM models. These results demonstrate that ADFSA enhances classification accuracy, improving training speed, lowering the inference latency, and minimizing memory usage. All mentioned characteristics make the proposed ADFSA suitable for scalable, real-time HAR applications on resource-constrained devices. The proposed technique effectively balances model complexity and predictive performance, validating its potential as a robust FS framework for video-based HAR systems.

\textbf{Future work}  will focus on leveraging self-supervised learning techniques for human activity recognition to minimize dependence on labeled datasets. Transformer-based temporal modeling will be investigated as a potential replacement for LSTM networks to enhance the capture of long-range motion dependencies. The proposed framework will also be adapted for deployment on low-power edge devices, such as the Raspberry Pi, to evaluate its real-time performance in resource-constrained settings. Furthermore, the methodology will be validated on more diverse and challenging video datasets, including HMDB51, to rigorously assess its generalization across a broad spectrum of activity categories and environmental conditions.

\end{document}